\documentclass{article}

\usepackage{arxiv}
\usepackage[utf8]{inputenc} 
\usepackage[T1]{fontenc}    
\usepackage{hyperref}       
\usepackage{url}            
\usepackage{booktabs}       
\usepackage{amsfonts}       
\usepackage{nicefrac}       
\usepackage{microtype}      
\usepackage{lipsum}
\usepackage{graphicx}
\graphicspath{ {./images/} }

\usepackage{mathtools}       
\usepackage{amssymb}         
\usepackage{amsthm}          
\usepackage{siunitx}         
\usepackage{bm}              
\usepackage{authblk}         
\usepackage[inline]{enumitem}

\DeclarePairedDelimiter\abs{\lvert}{\rvert}

\usepackage{caption}
\usepackage{subcaption}
\usepackage{makecell} 
\usepackage{multirow, multicol}

\usepackage[super,sort&compress,comma]{natbib} 
\usepackage[capitalize]{cleveref}
\crefformat{equation}{eqn~(#2#1#3)}
\crefrangeformat{equation}{eqns~(#3#1#4) to (#5#2#6)}

\newcommand{\angp}[1]{\phi^{err}_{#1}}
\newcommand{\magp}[1]{\lVert\mathbf{C}_\texttt{L}\rVert^{err}_{#1}}

\title{Geometry-free prediction of inertial lift forces in microfluidic devices using deep learning} 

\author[a, $\dagger$]{Jesse Ward-Bond}
\author[a, b, $\dagger$]{Ali Mashadian}
\author[a]{Timothy C. Y. Chan}
\author[a, c, d,  \thanks{Corresponding author: edmond.young@utoronto.ca}]{Edmond W. K. Young}

\affil[a]{Department of Mechanical \& Industrial Engineering, University of Toronto}
\affil[b]{Mechanical Engineering, Purdue University}
\affil[c]{Department of Materials Science \& Engineering, University of Toronto}
\affil[d]{Institute of Biomedical Engineering, University of Toronto}
\affil[$\dagger$]{These authors contributed equally to this work.}

\begin{document}
\maketitle
\begin{abstract}
Inertial microfluidic devices (IMDs) offer low-cost, high-throughput alternative techniques for many traditional particle- (or cell-) manipulation tasks, but simulating them requires being able to predict particle migration, and thus particle lift forces, under a variety of possible channel geometries. Recent work has demonstrated that machine learning models can be used to drastically speed up these numerical simulations, but doing so required training individual models for every unique channel cross-section type (e.g., rectangular, triangular)---shifting the burden from the simulation step to the training step. In this paper, we develop a novel approach for predicting particle lift forces that contains no explicit geometric parameters. We train a  neural network model using a new parameter set and show that while it performs comparably to existing models on channel geometries in the training set, it is able to generalize to unseen channel geometries far more effectively. We show that the lift force model developed herein can be easily transferred to particle tracing simulation software, where it is capable of predicting particle migration patterns consistent with the literature across a variety of channel designs.
\end{abstract}


\section{Introduction}
    
    Microfluidic technologies have advanced rapidly in the past two decades and are increasingly used for the transport and manipulation of suspended particles within microscale channels. Among these approaches, inertial microfluidics has emerged as a powerful passive technique for high-throughput particle and cell separation by exploiting inertial lift forces and secondary flow effects at intermediate Reynolds numbers.\cite{zhang2016} Current applications include cell sorting, circulating tumor cell enrichment, blood fractionation, and size-based nanoparticle purification.\cite{xiang2022, Lee2011, Zhou2013, Hur2012, Yoon2009, Guan2013, Park2009, Zhang2014, Bazaz2020} Future applications are expected to expand toward point-of-care diagnostics, continuous bioprocessing, and integrated sample preparation in automated lab-on-a-chip systems.\cite{jiang2022, jeon2022, volpe2019} Inertial microfluidic separation enables label-free concentration of target particulate species, narrowing of particle size distributions, and continuous operation without external fields or complex actuation, making it particularly attractive for scalable biomedical and industrial workflows.
    
    Most inertial microfluidic devices (IMDs) employ channels with rectangular cross-sections, largely reflecting legacy fabrication approaches such as soft lithography of poly(dimethylsiloxane) (PDMS), dry etching of glass, and micromilling of thermoplastics.\cite{sia2003, guckenberger2015, hwang2019} These fabrication methods naturally produce vertical sidewalls and planar geometries, which have historically constrained channel design despite the strong dependence of inertial focusing behaviour on channel cross-sectional shape. Significant research effort has therefore focused on optimizing flow conditions within rectangular geometries, as well as minimizing fabrication artifacts such as unintended sidewall bevelling that produces trapezoidal cross-sections and modifies equilibrium particle focusing positions. Alternative low-cost fabrication approaches, including thermally shrunk polymer methods (``Shrinky Dink''), have enabled semi-circular channel geometries; however, these techniques have generally suffered from limited resolution, poor surface smoothness, and reduced reproducibility, restricting their adoption for precision inertial flow studies.\cite{grimes2008}
    
    Recent advances in high-resolution additive manufacturing are beginning to transform microfluidic fabrication by enabling complex three-dimensional architectures that were previously inaccessible using conventional methods. Modern 3D printing technologies now permit fabrication of non-rectangular cross-sections with improved surface finish, optical transparency, and microscale feature resolution. Because inertial particle migration arises from the interplay between shear-gradient lift forces, wall-induced lift forces, and geometry-dependent secondary flows, channel cross-sectional shape represents a critical yet underexplored design parameter. Leveraging emerging fabrication capabilities provides an opportunity to move beyond traditional rectangular channels, expand the geometric design space, and engineer flow fields that enhance particle focusing stability, separation resolution, and throughput.\cite{tang2020} Such advances may enable the next generation of IMDs optimized through geometry-driven control of particle dynamics. 

    \begin{figure*}[!h]
    \centering
         \vspace{10pt}
         \includegraphics[alt={Five-panel flowchart: (a) 3D renders of rectangular, triangular, semicircular, and trapezoidal channels; (b) a heat map of velocity and a vector field of ground-truth lift forces, labeled with nondimensional parameters; (c) a four-layer FCNN architecture diagram followed by two scatter plots comparing DNS results to NN predictions for lift force components; (d) a software pipeline showing the model moving from PyTorch through ONNX to MATLAB and COMSOL; (e) 3D visualizations of particle focusing in square, cylindrical, and serpentine channels.}, width=\textwidth]{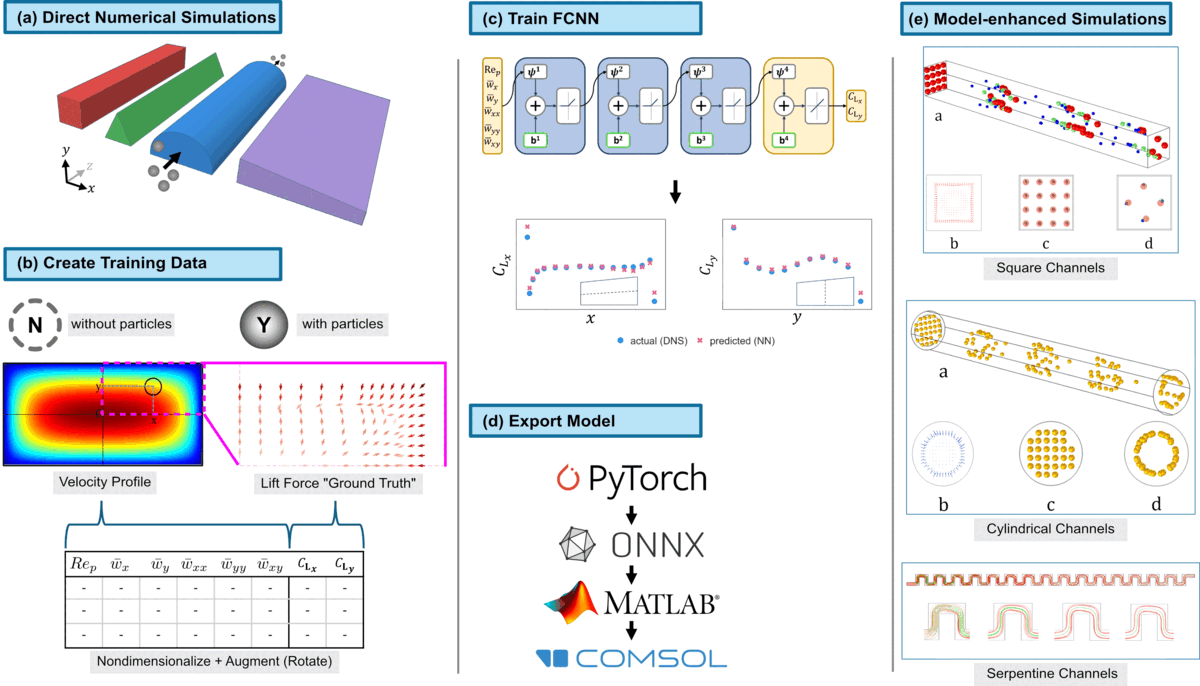}
         \caption[Figure abstract]{Overview of development and downstream integration of a geometry-free lift force model. (a) Direct numerical simulations (DNSs) are conducted to generate velocity profiles and lift force maps. (b) Results of the simulations are transformed according to our parametrization and augmented to be used as training data. (c) A fully connected neural network (FCNN) is trained to predict the x- and y-components of the lift force. (d) The model is exported for use in simulation software. (e) Particle-tracing simulations are conducted using model-predicted lift forces.}
         \label{f-abstract}
    \end{figure*}

    Fundamentally, the tendency of suspended particles flowing through a conduit to migrate to certain equilibrium positions, which was first observed experimentally at the macroscale by \citet{Segre1961} and then later observed by \citet{DiCarlo2007} at the microscale, depends on the net lift force acting on each particle. Despite advances in both theory and modelling technologies, it remains challenging to predict the net lift force in a computationally cheap and geometrically flexible manner.

    Researchers have attempted to derive analytical expressions for the lift forces acting on particles in IMDs, mostly through asymptotic analysis.\cite{Bazaz2020} However, these approaches rely on the limiting assumptions that particle diameters and Reynolds numbers ($Re$) are asymptotically small. Furthermore, these expressions are developed for particular channel geometries, and are not always readily adaptable. Although much work has been done to extend analytical approaches to more realistic microfluidic conditions,\cite{Hood2015, Asmolov2018, Su2023} these limitations persist.

    Given the challenges with analytical and semi-analytical approaches, IMD design typically relies on either physical design iterations or direct numerical simulations (DNSs). The former approach involves designing and fabricating the device, testing it on the bench, and repeating this process by trial and error until an optimal combination of design and operating characteristics emerges. This slow and stochastic process has spurred research into numerical simulation techniques that can predict the inertial migration of particles \textit{in silico}.\cite{DiCarlo2009b, Li2015, Liu2015, Shamloo2018} One common simulation workflow involves generating a map of lift forces by repeatedly fixing a stationary particle at different locations within a channel cross-section, allowing it to freely rotate and interact with the fluid, and then solving the underlying force balance to determine the net lift force. Once the lift force map is obtained, Lagrangian particle-tracing simulations can be used to determine particle migration.\cite{Liu2015, Shamloo2018} Although simulation results have shown great agreement with experimental results, they are computationally intensive and need to be re-run when design conditions change, making this workflow far from optimal.

    Advances in machine learning (ML) have been steadily percolating into microfluidic design workflows,\cite{Mcintyre2022, Owen2024} and some attempts have been made at accelerating IMD design with ML. In \citet{Su2021}, the authors trained multiple fully connected neural networks (FCNN) to generate lift force maps for straight channels with rectangular, semicircular, or triangular cross-sections, and then demonstrated that these maps closely agree with results from both physical experiments and DNSs. \citet{Hamdi2022} later extended that work by combining the neural networks for each cross section into a single neural network where cross-section shape was represented with a one-hot encoded vector and used as an input to their neural network. \citet{Safari2024} and \citet{Akhbari2024} trained FCNN and log-log regression models, respectively, to predict final particle positions in spiral channels, and were able to obtain lift coefficients and particle positions consistent with literature.\cite{Safari2024, Akhbari2024} These approaches demonstrate the feasibility of ML-assisted particle migration prediction, but have all resulted in models with explicit geometric dependence: they are either developed for particular channel cross-section shapes, or depend on parameters such as the channel's radius of curvature, which are not defined for many types of IMDs.

    In this paper, we introduce the first geometry-agnostic ML model for prediction of lift forces of particles travelling in microfluidic channels. We develop a parametrization scheme that allows this prediction to be independent of references to channel geometry and particle location. We show that the models developed using our parametrization scheme can generalize across diverse channel geometries that were not present in the training data. We further demonstrate that these models can be easily integrated into \texttt{COMSOL} and used to accurately predict particle migration in IMDs.

\section{Methods}
\label{methods}

    \Cref{f-abstract} gives an overview of the methodology used in this study. First, DNSs were conducted with and without particle tracing to obtain ``undisturbed'' (particle-free) velocity profiles and lift force maps for four different channel cross-sectional shapes under various flow conditions (\cref{f-abstract}a). This data was augmented (\cref{f-abstract}b) and a portion of this data was used to train and validate a FCNN model (\cref{f-abstract}c). The trained model was then exported to \texttt{COMSOL} MultiPhysics (\cref{f-abstract}d) and used to simulate different IMDs (\cref{f-abstract}e). Note that for straight channels, we used the convention that channel cross-sections are in the $xy$-plane, and that the channels extend longitudinally in the $z$-direction.

    \subsection{Geometry-agnostic formulation}
    In this section, we first identify the geometry-dependent parameters in a common lift force formulation. We then replace these parameters directly with the local velocity profile around the particle, which we model using a second-order Taylor series expansion. Finally, we non-dimensionalize our set of geometry-free parameters and use these as inputs to an FCNN.

    The lift force, $\mathbf{F}_\texttt{L}$, on a neutrally buoyant particle in a straight rectangular channel is commonly formulated as
    \begin{equation}
        \label{e-rect-principles}
        \mathbf{F}_\texttt{L} = f \left( \frac{a}{H}, \frac{x_0}{H}, \frac{y_0}{H}, AR, Re_c \right),
    \end{equation}
    where $a$ is the particle diameter, $H$ is the channel height, $(x_0,y_0)$ is the position of the center of the particle within the channel cross-section, $AR$ is the aspect ratio of the channel, and $Re_c$ is the channel Reynolds number.\cite{Liu2015, Su2021}. 

    Models developed from formulations like \cref{e-rect-principles} face two major challenges: 
    \begin{enumerate*}[label=(\alph*)]
        \item they depend on geometric characteristics that may not be defined for every channel type (e.g., aspect ratio is not well-defined for a trapezoid), and
        \item downstream simulations using \cref{e-rect-principles} will need to constantly determine channel cross-section geometries and the relative position of particles within those geometries, which may not be feasible in all channel geometries (e.g., expansion-contraction channels,\cite{Lee2011} or serpentine channels\cite{Zhang2014}).
    \end{enumerate*}

\begin{figure}
    \centering
        \includegraphics[alt={A schematic diagram illustrating a four-step transitive process: (1) An undisturbed velocity profile in a rectangular channel with laminar streamlines and a red parabolic vector overlay; (2) plus a spherical particle; (3) results in a disturbed velocity profile showing streamlines curving around the sphere and a green wake of secondary flow; (4) which corresponds to a spherical heat map of fluid stress on the particle surface, terminating in an arrow pointing to the lift force variable F_L.}, width=0.48\textwidth]{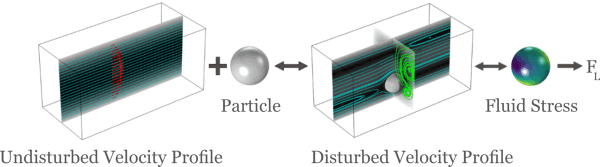}
        \caption{The transitive relationship between the undisturbed velocity profile and the lift force $\mathbf{F}_\texttt{L}$ acting on a particle.}
        \label{f-transitive}
    \end{figure}

    To address these challenges, we developed a novel model parametrization that does not explicitly depend on channel geometries. To do this, we first note that these geometric terms are only necessary insofar as they ultimately help specify the local velocity field, and thus the fluid-induced stresses, around a particle at a given position in the channel cross-section. We further note the following transitive relationships: the inertial lift force acting on a particle is determined by the fluid-induced stresses, which is in turn determined by the local velocity field, which is ultimately a function of the shape of the undisturbed velocity profile and the fluid and particle characteristics (\cref{f-transitive}). We exploit these transitive relationships to model the lift force acting on a particle as:
        \begin{equation}
            \label{e-no-geom}
            \mathbf{F}_\texttt{L} = f(\rho, \mu, a, \mathcal{U})
        \end{equation}
    where $\rho$ is the fluid density, $\mu$ is the dynamic viscosity, and
        \begin{equation}\label{e-surface-profile}
            \mathcal{U} := \left\{\mathbf{u}(\mathbf{x}) \;\middle|\; \|\mathbf{x} - \mathbf{x_0}\|_2 = \frac{a}{2}  \right\}
        \end{equation}
    is the set of velocity vectors $\mathbf{u}$ at every point on the surface of the particle, centred at $\mathbf{x_0}$, in the absence of fluid-particle interactions (i.e., using the undisturbed velocity profile).
    
    Models are only trained in straight channels containing Poiseuille flow, and thus velocity vectors in the undisturbed flow are only in the $z$-direction (i.e., the flow is perfectly perpendicular to the cross-section). Furthermore, the inertial lift force is zero in the $z$-direction. We can therefore simplify \cref{e-no-geom} to the two-dimensional case, only considering the velocities in the $z$-direction:
        \begin{equation}\label{e-no-geom2D}
            \mathbf{F}_{\texttt{L}_{XY}} \approx f(\rho, \mu, a, \mathcal{W}) \\
        \end{equation}
    where $w(x, y)$ is the $z$-direction velocity at point $(x,y)$ on the surface of the imaginary particle and
        \begin{equation}
            \mathcal{W} := \left\{w(x, y) \;\middle|\; (x - x_0)^2 + (y-y_0)^2 = \left( \frac{a}{2} \right)^2 \right\}
        \end{equation}
    is the set of all velocities in the $z$-direction, $w$, around the circumference of that particle in the $xy$-plane.

    Most particle-tracing simulations treat particles as point masses, and thus $\mathcal{W}$ is not computable. Instead, $\mathcal{W}$ can be approximated with the second-order Taylor series expansion around the particle centre (see Supplementary Information), which leads to the formulation:
        \begin{equation}
        \label{e-taylor}
            \mathbf{F}_{\texttt{L}_{XY}} \approx f(\rho, \mu, a, w, w_x, w_y, w_{xx}, w_{yy}, w_{xy})
        \end{equation}
    where $w_x$ and $w_y$ are first-order partial derivatives of the velocity field at the centre of the particle, and $w_{xx}$, $w_{xy}$ and $w_{yy}$ are the second-order derivatives.
    
    With the reformulation in \cref{e-taylor}, there is no longer any explicit dependence on channel geometries, and all model inputs are computationally tractable. All that remains is to non-dimensionalize the formulation using the Buckingham~$\pi$ theorem.\cite{Buckingham1914} This results in a final expression for the dimensionless lift coefficient, $\mathbf{C}_{\texttt{L}_{XY}}$, given as:
        \begin{equation}
            \label{e-no-dim} 
            \mathbf{C}_{\texttt{L}_{XY}}  = \frac{\mathbf{F}_{\texttt{L}_{XY}}}{\rho w^2 a^2} = 
            f\left(\frac{\rho w a}{\mu}, 
            \frac{w_xa}{w}, \frac{w_ya}{w},
            \frac{w_{xx}a^2}{w}, \frac{w_{yy}a^2}{w}, \frac{w_{xy}a^2}{w}\right)
        \end{equation}
    For convenience, we use the following shorthand to represent model \cref{e-no-dim} going forward:
        \begin{equation}
            \label{e-no-dim2}
            \mathbf{C}_\texttt{L}= f(Re_p, \bar{w}_x, \bar{w}_y, \bar{w}_{xx}, \bar{w}_{yy}, \bar{w}_{xy}).
        \end{equation}
    where the overbar indicates non-dimensionalized parameters. Note that we only consider lift force in the $xy$-plane for the remainder of this study.

    \subsection{Data generation}
\begin{figure*}
    \begin{center}
    \begin{subfigure}{\textwidth}
        \centering
        \includegraphics[alt={Four geometric cross-sections plotted on xy-coordinate planes with vertices labeled: (1) a red square with a black star at the origin (0,0) and a vertex at (25,25), featuring dashed lines along the axes; (2) a green right triangle with vertices at (0,-25), (25,-25), and (0,25), with a vertical dashed line and a star on the y-axis; (3) a blue quadrant (quarter-circle) with vertices at (0,-25) and (50,-25), a dashed vertical line, and a star on the y-axis; (4) a purple trapezoid with vertices at (-300,0), (300,0), (300,150), and (-300,80), with a star located at the midpoint of the base. An xyz coordinate legend indicates x is horizontal, y is vertical, and z points into the page.}, width=\textwidth]{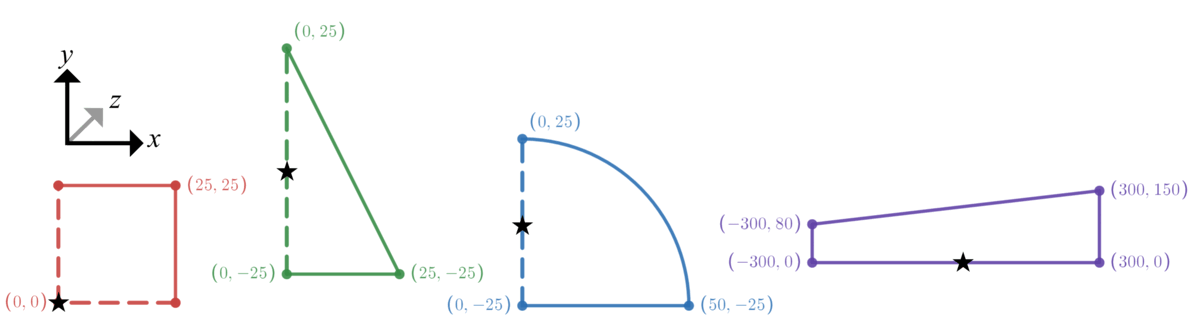}
    \end{subfigure}
    \begin{subfigure}{\textwidth}
        \centering\small
          \begin{tabular*}{\textwidth}{@{\extracolsep{\fill}}cccccc}
            \hline
            \textbf{Geometry} & \textbf{AR}   & \textbf{a} (\unit{\micro\meter})               & $U_{m}$ (\unit{\meter\per\second}) & \textbf{Size} & \textbf{Source}\\
            \hline
            $\mathcal{R}$   & $\{1,2,3,4\}$ & $\{5.0, 7.5, 10.0, 12.5, 15.0\}$  & $\{1.0, 2.0, 3.0, 4.0\}$ & 8068 & \cite{Su2021}\\
            $\mathcal{T}$   &  -           & $\{5.0, 7.5, 10.0, 12.5, 15.0\}$  & $\{1.0, 2.0, 3.0, 4.0\}$ & 2420 & \cite{Su2021}\\
            $\mathcal{S}$   &  -           & $\{5.0, 7.5, 10.0, 12.5, 15.0\}$  & $\{1.0, 2.0, 3.0, 4.0\}$ & 3672 & \cite{Su2021}\\
            $\mathcal{P}$   &  -           & $\{10.0, 12.5, 25.0\}$            & $\{0.25, 0.52, 0.63, 1.3, 2.0,  2.6, 4.2\}$& 1408 & This paper\\
            \hline
          \end{tabular*}
    \end{subfigure}
    \end{center}
    \vspace{-3mm}
    \caption[Dataset summary]{Summary of raw datasets used for model training and testing. From left to right: rectangular ($\mathcal{R}$), triangular ($\mathcal{T}$), semicircular ($\mathcal{S}$), and trapezoidal ($\mathcal{P}$). Solid lines indicate channel walls and dashed lines are lines of symmetry and also indicate the geometric limits of the simulated data. Table abbreviations: aspect ratio (AR), particle size (a), maximum flow velocities ($U_m$), cardinality (size), and source of each dataset. Note that we did not use all combinations of particle sizes and maximum flow velocities for the trapezoidal cross-sections.}
    \label{f-cross-sections}
    \end{figure*}

    Lift force maps for particles in rectangular ($\mathcal{R}$), triangular ($\mathcal{T}$), and semicircular ($\mathcal{S}$) channels were taken from existing datasets reported in literature. These datasets contain lift force maps for different particle sizes, maximum fluid velocities ($U_m$) and, for the rectangular channels only, aspect ratios, as outlined in \cref{f-cross-sections}.\cite{Su2021} An additional independently-generated trapezoidal dataset was created following the process in Shamloo \textit{et al.}\cite{Shamloo2018} Datasets were augmented by rotating in \qty{20}{\degree} increments and flipping about the $X$-axes, resulting in a 36-fold increase in the size of each dataset. Each dataset was split into training/validation/testing sets (0.7/0.15/0.15) denoted  $\mathcal{R}^{tr}$, $\mathcal{R}^{v}$, $\mathcal{R}^{ts}$, respectively.

    \subsection{Machine learning model}
\begin{figure}[h]
    \begin{center}
        \includegraphics[alt={A diagram of a neural network with four hidden layers.}, width=0.48\textwidth]{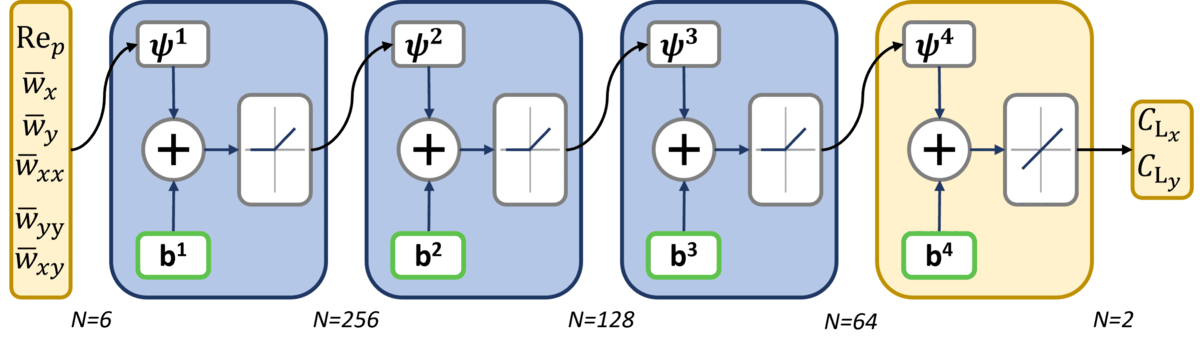}
    \end{center}
    \vspace{-3mm}
    \caption[Model architecture]{Model architecture. Input nodes are on the left, output nodes on the right. Hidden layers are represented in blue. Biases ($\text{b}$) are outlined in green. $N$ is the number of neurons per layer. $\psi$ are model weights.}
    \label{f-fcnn}
    \end{figure}

    A fully connected neural network (FCNN) was used to approximate \cref{e-no-dim2}. In these networks the input to a given neuron is a linear combination of the outputs from all neurons in the previous layer. Inputs were the arguments of \cref{e-no-dim2} and outputs were the $\mathbf{C}_\texttt{L}$ components: $(C_{\texttt{L}_x}, C_{\texttt{L}_y})$. The ReLU activation function was used for all layers except the final layer, which used linear activation functions. Models were trained using minibatch gradient descent with Nesterov Momentum for 300 epochs with early stopping if the validation loss did not improve over the incumbent after 30 epochs.\cite{Nesterov1983, sutskever2013} The loss function was
    \begin{equation}
        \label{e-mse-loss}
        \mathcal{L}(\mathcal{B}_t) = \frac{1}{\left| \mathcal{B}_t \right|} \sum_{i=1}^{\left| \mathcal{B}_t \right|} \left\| \mathbf{C}_{\texttt{L}}^i - \hat{\mathbf{C}}_{\texttt{L}}^i \right\|_2^2
    \end{equation}
    \noindent where $\mathcal{B}_t$ is batch $t$ and $\hat{C}_{\texttt{L}_x}^i$ is the predicted $x$-component of the lift coefficient vector for the $i^{th}$ data point in batch $\mathcal{B}_t$. Hyperparameter tuning was done on the validation datasets following the methodology outlined in \citet{Godbole2023} The final architecture was $6 \times 256 \times 128 \times 64 \times 2$ (\cref{f-fcnn}).

    In addition to mean squared error (MSE), we also evaluate models using the angular error of predictions, which we define as
        \begin{equation}
            \phi^{err}=\cos^{-1} \left( \frac{\mathbf{C}_\texttt{L}\cdot\hat{\mathbf{C}_\texttt{L}}}{\lVert\mathbf{C}_\texttt{L}\rVert\cdot\lVert\hat{\mathbf{C}_\texttt{L}}\rVert} \right),
        \end{equation}
    and the relative magnitude error of predictions, which we define as
        \begin{equation}
            \lVert\mathbf{C}_\texttt{L}\rVert^{err} = \abs{\frac{\lVert\hat{\mathbf{C}_\texttt{L}}\rVert-\lVert\mathbf{C}_\texttt{L}\rVert}{\lVert\mathbf{C}_\texttt{L}\rVert}} \times 100\%.
        \end{equation}
    Collectively, these two metrics provide a measure of the accuracy of the \textit{direction} and the \textit{magnitude} of the predicted lift force, respectively. 
    
    A baseline model was trained on $\mathcal{R}^{tr}$ in \texttt{MATLAB} according to the methodology developed by \citet{Su2021} This training used the same dataset, architecture, Levenberg–Marquardt loss function, and stopping conditions as reported by the original authors. To compare that model to our model, both  were trained and tested on the same unrotated, unflipped, rectangular dataset $\mathcal{R}_0 \subset \mathcal{R}$, which contained 5660 training points and 2328 testing points, a 70/30 split. To enable direct comparison, outputs from the baseline model were converted to match our $\mathbf{C}_\texttt{L}$ formulation (see Supplementary Information).
    
    Feature importance was estimated using Shapley values. Shapley values were in turn estimated using the \texttt{KernelExplainer} module from the \texttt{SHAP} library on 500 samples from $\mathcal{R}^{ts}$.\cite{lundberg2017}

\subsection{Downstream simulation}
\label{s-downstream}

    Trained models were converted into the Open Neural Network Exchange (ONNX) format.\cite{Bai2019} This format allowed the model to be imported into \texttt{MATLAB}, which in turn enabled it to be used for particle-tracing simulations within \texttt{COMSOL} Multiphysics. Particle tracing simulations were performed for both straight and curved channels. Straight channel geometries included channels with square, triangular, circular, trapezoidal, and rhombic cross-sections, as well as expansion-contraction channels with rectangular cross-sections. Curved channel geometries included the standard spiral and u-turn serpentine designs, as well as a ``reverse wavy'' serpentine geometry based on the work by \citet{Zhou2018}

    All the training data used in this paper were from straight microchannels, so the model was parametrized only with velocity gradients in the $XY$-direction (\cref{e-no-dim}). We used a simple rotational mapping to account for the non-zero $Z$-gradients in curved channels. Where necessary, velocity gradients in the three dimensional $XYZ$-space were converted to the $XY$-plane with a rotational mapping that aligned the velocity vector at the particle centre with the $Z$-direction. This mapping was obtained using Rodrigues' rotation formula.\cite{Dai2015} The predicted $\mathbf{C}_{\texttt{L}_{XY}}$ values were then transformed back into the correct direction using the inverse mapping. Full details can be found in Supplementary Information.

\section{Results}
\label{results}

    \subsection{Model performance}
\begin{figure}[h]
        \centering
        \begin{minipage}{0.51\textwidth}
            \centering
            \includegraphics[alt={Two side-by-side percentile plots comparing the Current Study (solid blue line) to Su et al. (dashed orange line). Both plots feature a logarithmic y-axis and a linear x-axis from 0 to 100 percentile. The left plot shows relative magnitude error in percent, with both models tracking closely between 0.1\% and 100\% for most of the distribution. The right plot shows angular error in degrees, with both curves following a similar upward trend between 0.001 and 1 degree. In both plots, the dashed orange line remains slightly below the solid blue line across the middle percentiles, indicating slightly lower errors for the model from Su et al. before both converge at the extreme high percentiles.}, width=\linewidth]{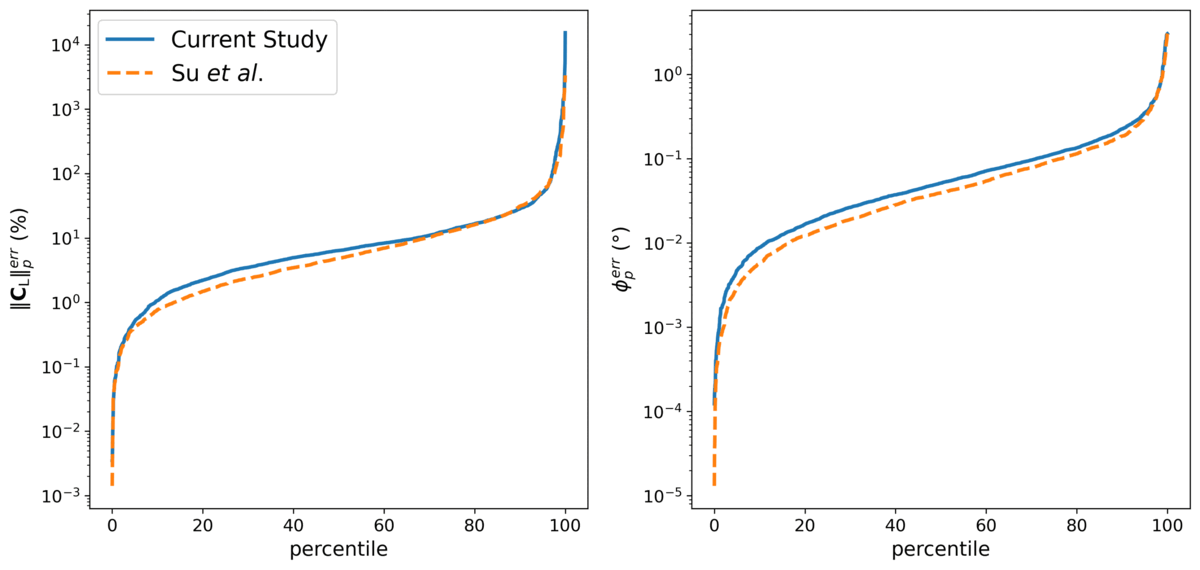}
        \end{minipage}
        \hfill
        \begin{minipage}{0.49\textwidth}
            \centering
            \small
            \renewcommand{\arraystretch}{1.25}
            \begin{tabular*}{\linewidth}{@{\extracolsep{\fill}}rcc}
                \toprule
                \textbf{Metric} & \textbf{Current Study} & \textbf{Su \textit{et al.}} \\
                \midrule
                \textbf{MSE} & \num{8.64e-7} & \num{9.93e-7} \\
                $\mathbf{\phi}^{err}_{50}$ (\unit{\degree}) & \num{2.9} & \num{2.2} \\
                $\lVert\mathbf{C}_\texttt{L}\rVert^{err}_{50}$ (\%) & \num{6.4} & \num{4.8} \\
                \boldmath$R^2_x$ & \num{0.995} & \num{0.995} \\
                \boldmath$R^2_y$ & \num{0.995} & \num{0.993} \\
                \bottomrule
            \end{tabular*}
        \end{minipage}
    \vspace{2mm}
    \caption[Model comparison]{Comparison between our model and the model developed in \citet{Su2021} Top: percentile plots of the relative magnitude error (left) and angle (right) between actual and predicted lift force coefficients. Bottom: summary statistics. $\mathbf{\phi}^{err}_{50}$ and $\lVert\mathbf{C}_\texttt{L}\rVert^{err}_{50}$ are the median angle and relative magnitude errors, respectively.} 
    \label{f-performance-comparison}
    \end{figure}

    \Cref{f-performance-comparison} shows the prediction performance of a model trained with our parametrization compared to one trained with the parametrization from \citet{Su2021}
    When both models are trained and tested on unrotated, unflipped datasets, we found that our model had a lower MSE than the model from \citet{Su2021}. Despite this lower error, our model was slightly worse at predicting both the magnitude and direction of $\mathbf{C}_\texttt{L}$ vectors. The median angle and magnitude errors for our model were \qty{2.9}{\degree} and \qty{6.4}{\percent} respectively, compared to \qty{2.2}{\degree} and \qty{4.8}{\percent} for the re-implemented Su model (\cref{f-performance-comparison}, inset). Percentile plots confirmed that the re-implemented model predicted magnitude and angle better for up to \qty{70}{\percent} and \qty{95}{\percent} of samples, respectively (\cref{f-performance-comparison}).

\begin{figure}
    \centering
        \includegraphics[alt={ Three-panel stacked plot showing error metrics for different test sets, grouped by color: Rectangular (red), Triangular (green), Semicircular (blue), and Trapezoidal (purple). Within each color group, four bars or boxplots represent training sets of increasing complexity, distinguished by textures: solid (Rectangular only), diagonal stripes (Rectangular and Triangular), cross-hatch (Rectangular, Triangular, and Semicircular), and dots (all four geometries). The top panel shows Mean Squared Error (MSE) on a logarithmic scale. Adding a geometry to the training set significantly drops its MSE in the corresponding test group. The middle panel shows boxplots for angular error in degrees, and the bottom panel shows boxplots for relative magnitude error in percent. Across all panels, adding geometries to the training set increases test performance on that same geometry, at the cost of decreasing test performance on existing geometries in the training dataset. The Trapezoidal test set (purple) maintains relatively high and consistent error across all training configurations compared to the other geometries}, width=0.8\textwidth]{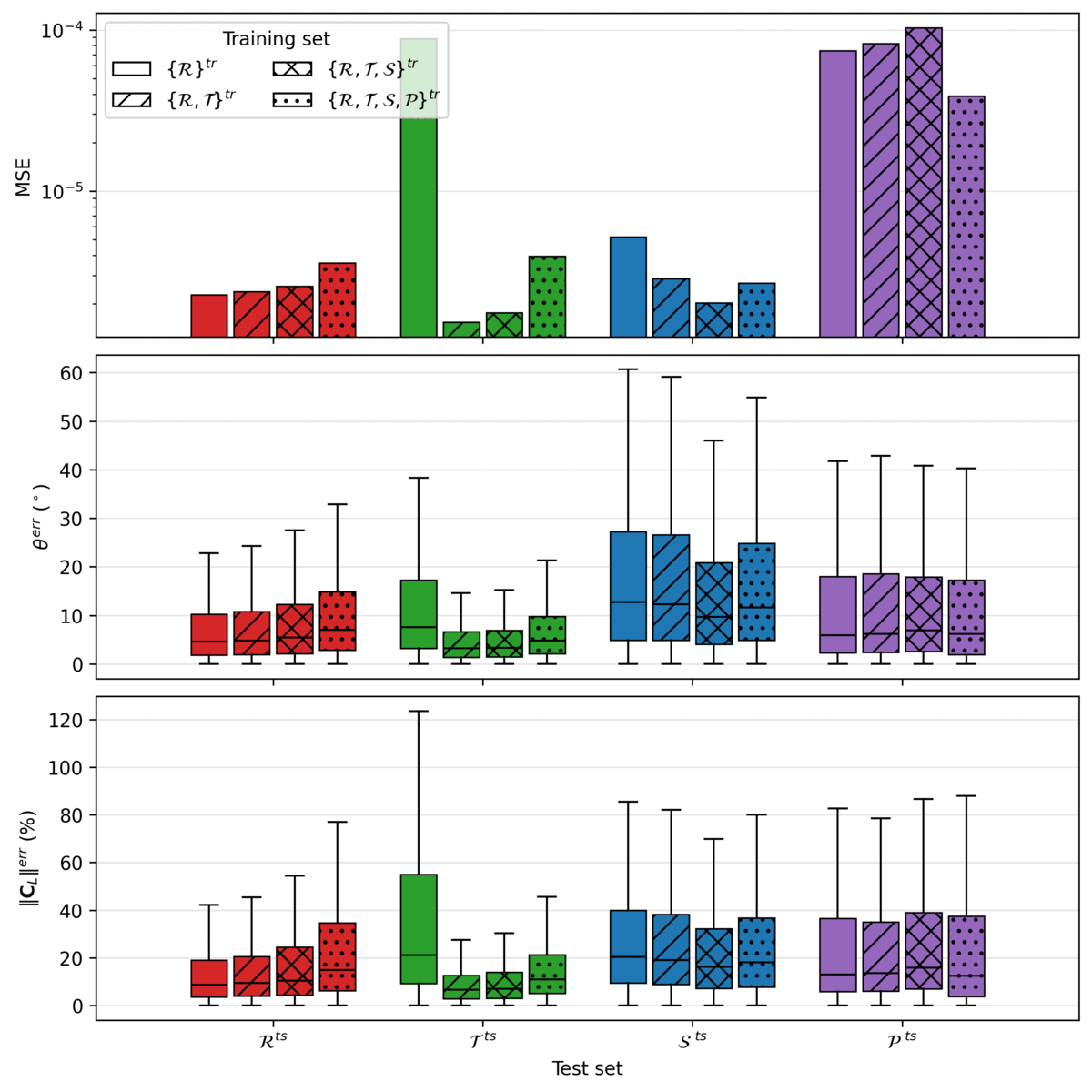}
    \caption[Generalization of the best trained model to other geometries]{Model performance on seen and unseen geometries as new geometries are added to the training data. Models were retrained completely when additional geometries were added to the training set. $\mathcal{R, T, S, P}$, represent the rectangular, triangular, semicircular, and trapezoidal datasets, respectively. Superscripts $tr$ and $ts$ represent training and testing subsets.}
    \label{f-generalization}
    \end{figure}

    To investigate the generalization performance of our model, we re-trained it several times with increasing amounts of data from new geometries (\cref{f-generalization}). When trained only on rectangular data, the model (predictably) performed best on the rectangular test set: \qty{50}{\percent} of predicted vectors are within \qty{4.6}{\degree} of their true directions, and within \qty{8.7}{\percent} of their true magnitude. When additional geometries were added to the training data, the model performed better on the corresponding test set for the new geometry, at the cost of some performance loss on the existing geometries. For example, when triangular training data was added, the median angle and magnitude errors ($\angp{50}$ and $\magp{50}$) on the triangular test data were \qty{57}{\percent} and \qty{69}{\percent} lower than they were when the model was only trained on rectangular data. At the same time, adding this data caused the $\angp{50}$ and $\magp{50}$ scores to increase for the rectangular test data by \qty{0.1}{\degree} and \qty{0.8}{\percent} respectively. There was a similar pattern when data from semicircular geometries were added to the training data, although in this case the increase in $\angp{50}$ and $\magp{50}$ scores across the other geometries was not as pronounced. Adding the trapezoidal training data resulted in the largest increase in $\angp{50}$ and $\magp{50}$ scores across all geometries, likely because this data contains flow rates and particle sizes that the other datasets do not.

    Across all the geometries, and for all permutations of training data tested in \cref{f-generalization}, the $\angp{50}$ and $\magp{50}$ errors were highest for the semicircular test data. The $90^{th}$ percentile angle and magnitude errors ($\angp{90}$ and $\magp{90}$) were highest in the trapezoidal data. This suggests that the trapezoidal test data has a wider distribution, which we confirmed with principal component analysis (see Supplementary Information).

    \Cref{f-meridians} shows a qualitative examination of the ability of our model to generalize to triangular, semicircular, and trapezoidal geometries when only trained on rectangular data. The model was able to capture much of the complex $\mathbf{C}_{\texttt{L}_x}$ and $\mathbf{C}_{\texttt{L}_y}$ variation within the cross sections, although it was slightly less faithful at capturing $\mathbf{C}_{\texttt{L}_x}$ variation across the $X$-direction than $\mathbf{C}_{\texttt{L}_y}$ variation across the $Y$-direction. These results were consistent across all flow rates and particle sizes in \cref{f-cross-sections}.

    \begin{figure}
    \centering
        \includegraphics[alt={Six rows of comparison plots for three channel geometries: triangular (top two rows), semicircular (middle two rows), and trapezoidal (bottom two rows). Each row consists of three panels: a heatmap of actual values, a heatmap of predicted values, and a scatter plot of a meridian cross-section. The heatmaps use a diverging blue-to-red color scale centered at zero. The first and second columns show that the neural network predictions qualitatively capture the spatial distribution of the CL_x and CL_y components across all three geometries. The third column contains scatter plots titled "Horiz. Meridian" or "Vert. Meridian," where blue dots (actual) and pink crosses (predicted) are plotted against coordinate position. These plots show close agreement in trends, though the predicted values exhibit more visible oscillation and noise for the CL_x values in the triangular and semicircular cases compared to the other figures. Insets in the scatter plots indicate the specific horizontal or vertical meridian line used to create the final figure in each row.}, width=0.6\textwidth]{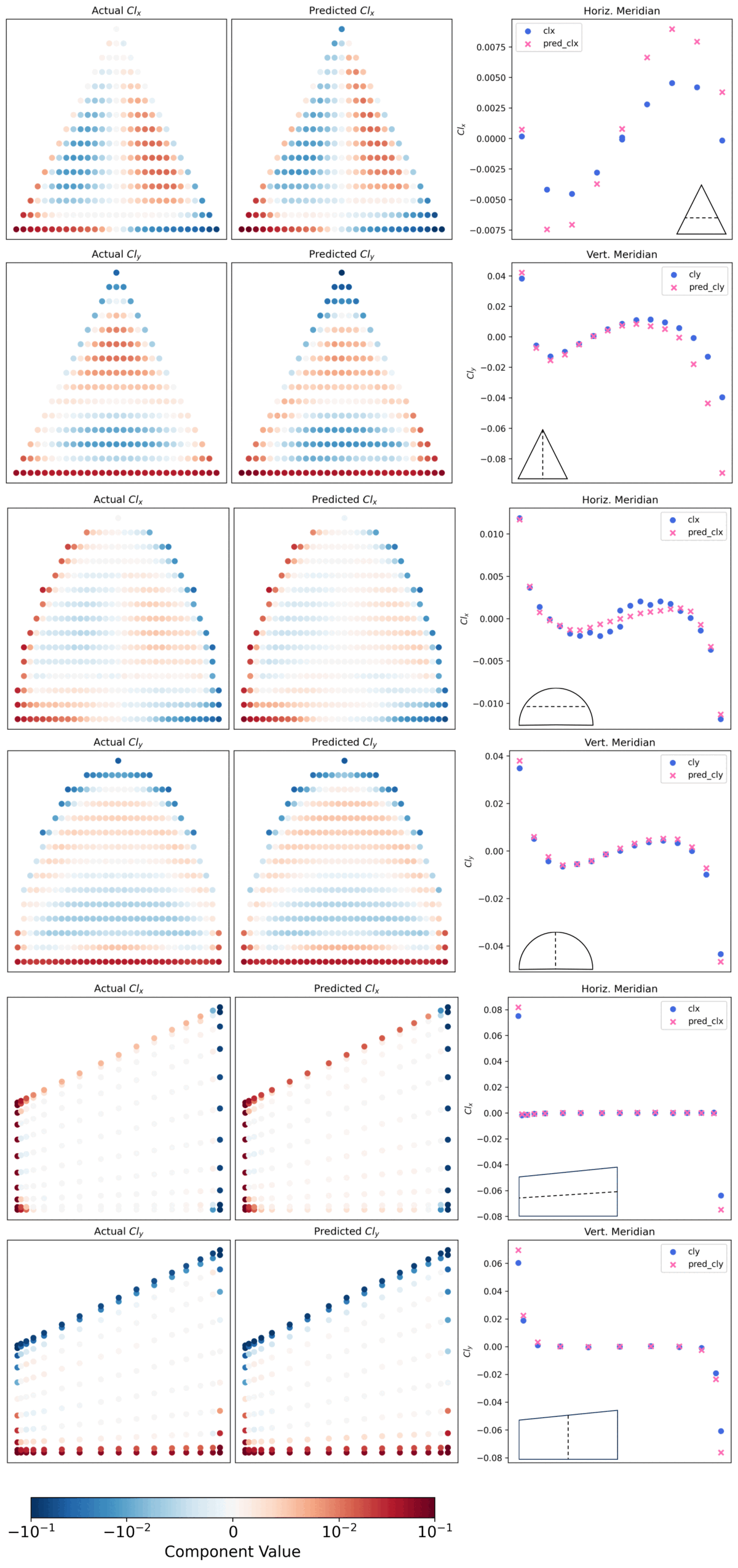}
    \caption[Meridian plots for different geometries]{Comparison of DNS- and FCNN-predicted $C_{\texttt{L}_x}$ and $C_{\texttt{L}_y}$ profiles in straight channels with different cross-sections. The model was trained only on rectangular cross sections. Channel dimensions are not to scale.}
    \label{f-meridians}
    \end{figure}

    \subsection{Downstream simulation}

\begin{figure*}
    \centering
    \includegraphics[alt={Four-panel simulation summary: (a) Particle tracing in a straight square channel where red, green, and blue particles of different sizes transition from a uniform grid at the inlet to four focused equilibrium positions at the outlet, shown with a corresponding lift force vector map; (b) tracing in a triangular channel where yellow particles migrate from a random distribution to three equilibrium positions near the midpoints of the channel walls; (c) An expansion-contraction simulation with two particle sizes. After 25 chambers, the smaller particles have migrated to two streamlines closer to the channel wall, while the large particles have stayed in the middle. This is presented alongside experimental results from literature validating this migration pattern. (d) As in (c), but in a serpentine channel. The particles exhibit the same migration pattern.}, width=\textwidth]{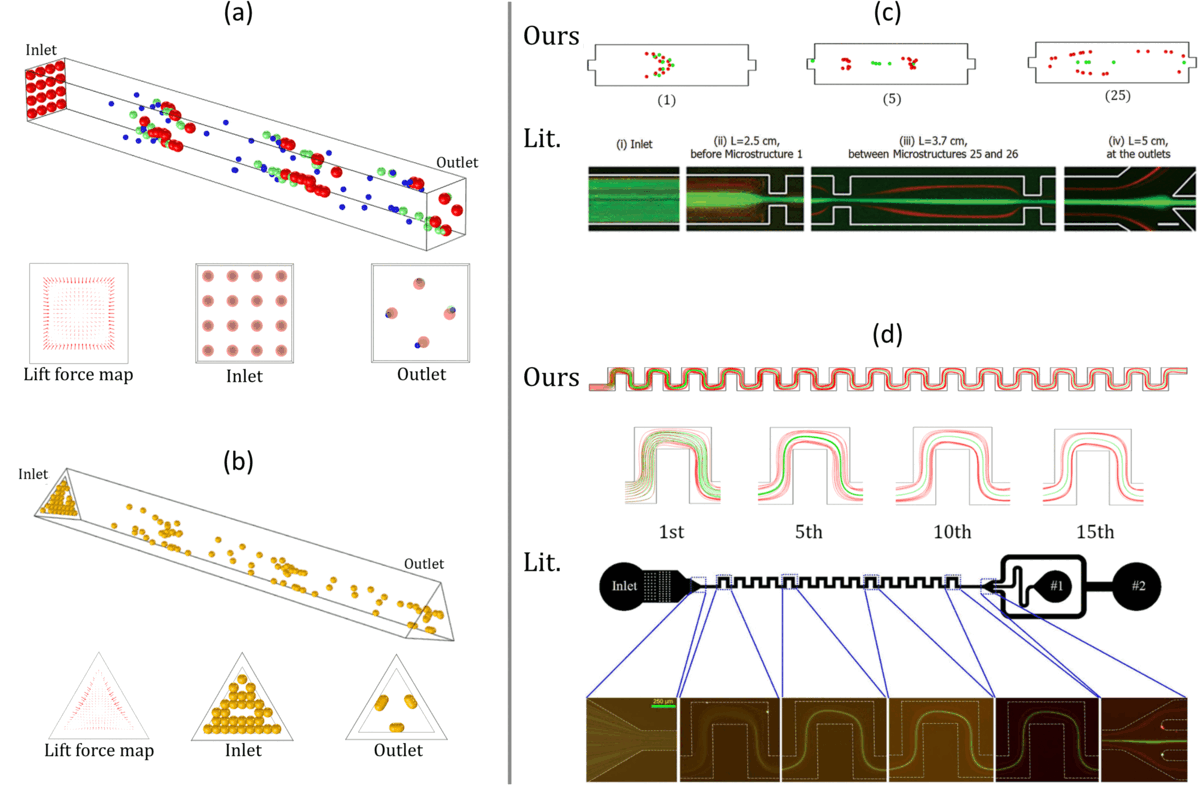}
    \caption[Downstream simulations]{Particle trajectories from \texttt{COMSOL} simulations employing our lift-force model trained on straight channels with rectangular, semicircular, triangular, and trapezoidal cross-sections. (a) A straight square channel containing \qty{10}{\micro\meter} (blue), \qty{15}{\micro\meter} (green), and \qty{20}{\micro\meter} (red) particles. (b) A straight triangular channel containing \qty{10}{\micro\meter} particles. (c) Twenty-five chambers of an expansion-contraction channel containing 5.5 (red) and \qty{9.9}{\micro\meter} particles. (d) A 15-unit u-turn serpentine channel containing \qty{3}{\micro\meter} (red) and \qty{10}{\micro\meter} (green) particles. Experimental results from literature for (c) were adapted from \citet{Wu2016} with permission from the Royal Society of Chemistry. Experimental results from literature for (d) were adapted from \citet{Zhang2014} with permission from Springer Nature.}
    \label{f-simulation}
    \end{figure*}

    Straight channel simulations included square, triangular, circular, trapezoidal, and rhomboidal cross-sections. \Cref{f-simulation}a and \cref{f-simulation}b show the results for rectangular and trapezoidal cross-sections. In all cases, the simulated particles migrated to focal points matching those reported in the literature. For example, \cref{f-simulation}a illustrates a particle-tracing simulation in a \qty{100}{\micro\meter} $\times$ \qty{100}{\micro\meter} $\times$ \qty{5}{\centi\meter} rectangular channel. Particles of varying sizes were released uniformly at the inlet and migrated to the four equilibrium positions observed in the literature. Similarly, \cref{f-simulation}b shows \qty{10}{\micro\meter} particles focused to the expected three positions in a triangular channel with a \qty{100}{\micro\meter} base. We additionally obtained particle focusing results consistent with the literature for circular, trapezoidal, and rhombic cross-sections (\cref{s-supp-sim}).

    We further tested the performance of our model in particle tracing simulations in curved channels, channels with non-constant cross sections, and channels with turns. \Cref{f-simulation}c shows the results of simulating particle tracing for \qty{5.5}{\micro\meter} and \qty{9.9}{\micro\meter} particles flowing through an expansion-contraction channel. Our simulation predicted that, after 25 chambers, the smaller particles fully migrate to streamlines closer to the channel walls, while the larger particles focus on the channel centerline. These simulation results are closely aligned with experimental findings in existing literature.\cite{Wu2016} \Cref{f-simulation}d shows similar agreement between a simulated serpentine channel using our lift force model and experimental observations from literature: the smaller particles (\qty{3}{\micro\meter}, red) migrated to streamlines closer to channel walls, while the larger particles (\qty{10}{\micro\meter}, green) focused along the channel centerline.\cite{Zhang2014} Additional comparisons between simulations and experimental observations for ``reverse-wavy'' and spiral IMDs are available in the Supplementary Information.
    
    \subsection{Feature importance}
\begin{figure}[h]
    \begin{center}
        \includegraphics[alt={Two vertically stacked SHAP summary plots showing feature impact on model output for x and y lift force components. Features are listed on the y-axis, SHAP values on the x-axis, and color represents feature value from low (blue) to high (pink). In the top plot, w_x is the most influential feature, where high values (pink) correlate with positive SHAP values and low values (blue) with negative ones. In the bottom plot, w_y is most influential with a similar trend. In both plots, the Reynolds number (Re_p) and higher-order velocity derivatives are clustered tightly around the zero line, indicating they have significantly less impact on the model predictions than the primary velocity components.}, width=0.48\textwidth]{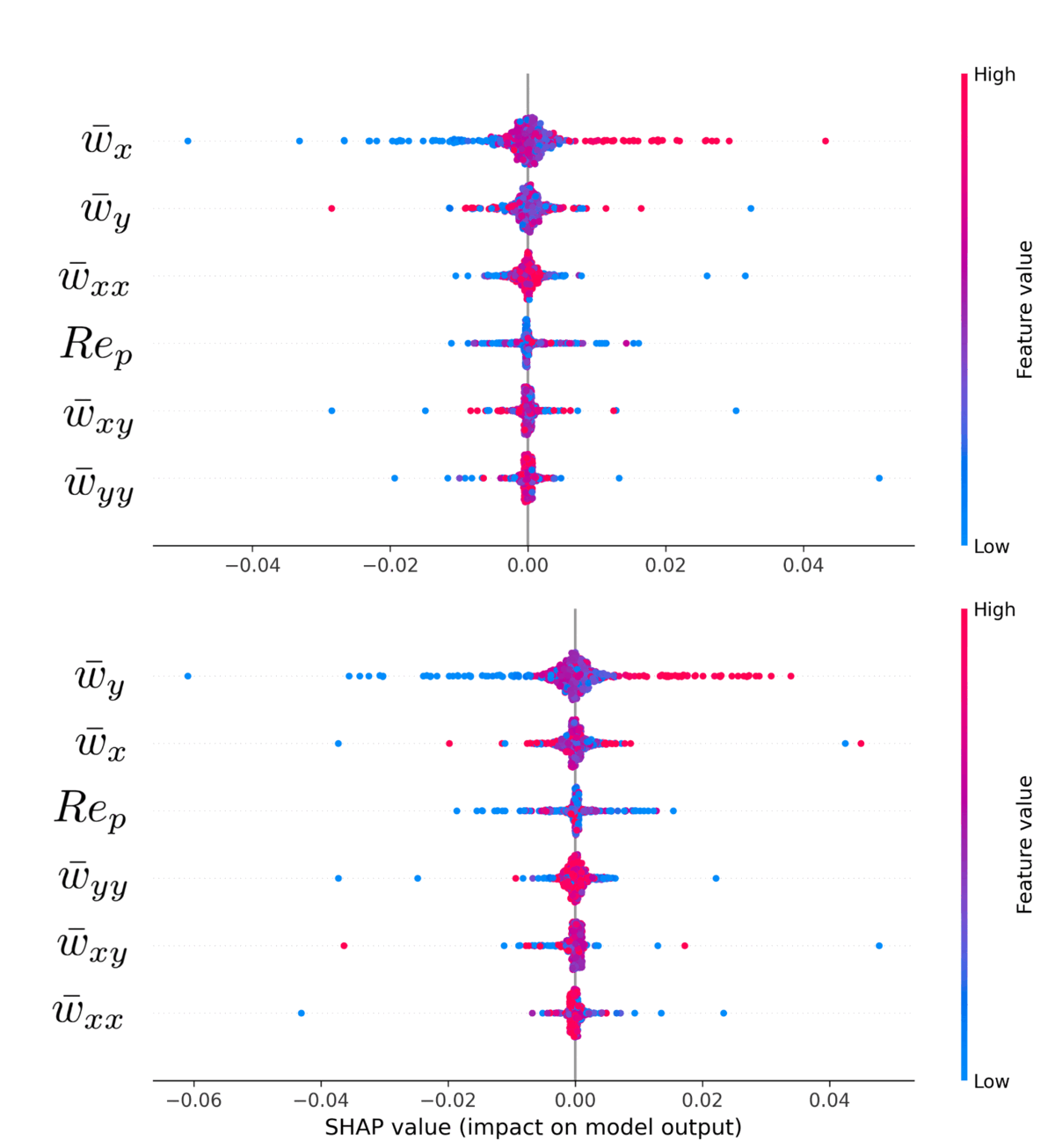}
    \end{center}
    \vspace{-3mm}
    \caption[Shapley values for a model trained on rectangular data]{Shapley values for $C_{\texttt{L}_x}$ (top) and $C_{\texttt{L}_y}$ (bottom), calculated using the \texttt{SHAP} library for python on 500 samples from the rectangular test data $\mathcal{R}^{ts}$.}
    \label{4-shap}
    \end{figure}

    Feature importance analysis showed that $\bar{w}_x$ and $\bar{w}_y$ had the greatest impact on $C_{\texttt{L}_x}$ and $C_{\texttt{L}_y}$, respectively. As expected from the shape of the velocity profile and the nature of wall-induced lift forces (\cref{4-shap}), a larger magnitude in the first derivative term (i.e., a steeper profile) led to a larger contribution to the corresponding $\mathbf{C}_{\texttt{L}}$ component.\cite{lundberg2017} The remaining features had less pronounced effects on model outputs and behaved in less predictable ways. It is possible that our current parametrization does not fully represent all variance in the lift force. Additional features such as hydraulic diameter or a modified particle blockage ratio defined relative to cross-sectional area rather than a single characteristic dimension may improve model performance, and both can be defined even for complex geometries.

    \section{Discussion}
    \label{discussion}

    \subsection{Model performance}
    Even when trained on rectangular data alone, our model was still able to provide reasonable predictions of $\mathbf{C}_\texttt{L}$ vectors in the other geometries. This is qualitatively confirmed in \cref{f-meridians}, where the model was only trained on rectangular geometries but successfully learned much of the behaviour of $\mathbf{C}_{\texttt{L}_x}$ and $\mathbf{C}_{\texttt{L}_y}$ within the other channel geometries. These results, together with the downstream simulation results in \cref{s-downstream}, lead us to conclude that our parametrization can successfully predict lift force without requiring geometric parameters, and that this results in models which can generalize across geometries.

    An obvious caveat to the above claim is that it appears adding new geometries to the training data decreased performance on the existing geometries (\cref{f-generalization}). This suggests distributional shifts between the datasets. Adversarial validation experiments with and without target values indicated that this was \textit{data} drift and not \textit{concept} drift (\cref{f-adversarial}). In other words, the feature-target relationship (\cref{e-no-dim2}) stayed consistent between datasets, but the distributions of the features themselves changed. The data drift was particularly striking in the triangular and trapezoidal  datasets, which were also the hardest geometries for the rectangular-only model to predict (\cref{f-generalization}). It is difficult to determine if this input data drift is an artifact of simulation differences or a result of some unintended geometry dependence in our formulation (\cref{e-no-dim2}). For the trapezoidal data it seems likely that the distributional shift was a case of the former, as the dataset contained particle sizes, midline velocities, and channel heights that the other three datasets did not have.

    \subsection{Simulation in curved channels}
    Our model successfully predicted particle migration even in channels with curves, corners, and non-constant cross-sections. In such channels, particle migration is understood to depend partly on the development of secondary flows and the resulting Dean force.\cite{Nivedita2017} These forces were not present in the training data, which only contained data from straight channels. Additionally, our model was only trained to predict lift forces in cross-sectional directions (the $XY$-plane, in a straight channel), but particles in non-straight channels experience lift forces outside of the cross-sectional plane. It is noteworthy that, given these limitations, the combination of our lift force model and rotational mapping technique proved robust enough to accurately predict particle trajectories, even in the presence of the complex fluid dynamics found in non-straight channels.

    \subsection{Error profile}
\begin{figure}[h]
    \begin{center}
    \begin{subfigure}{0.4\textwidth}
        \caption{}
        \centering
        \includegraphics[alt={Two log-log scatter plots comparing predicted lift force coefficients to actual values for x and y components. Both plots show a dense cloud of blue data points clustered around a diagonal 1-to-1 dotted line ranging from 10^-9 to 10^-1. Prediction accuracy is highest at large magnitudes, with increased dispersion occurring for values below 10^-5.},width=\textwidth]{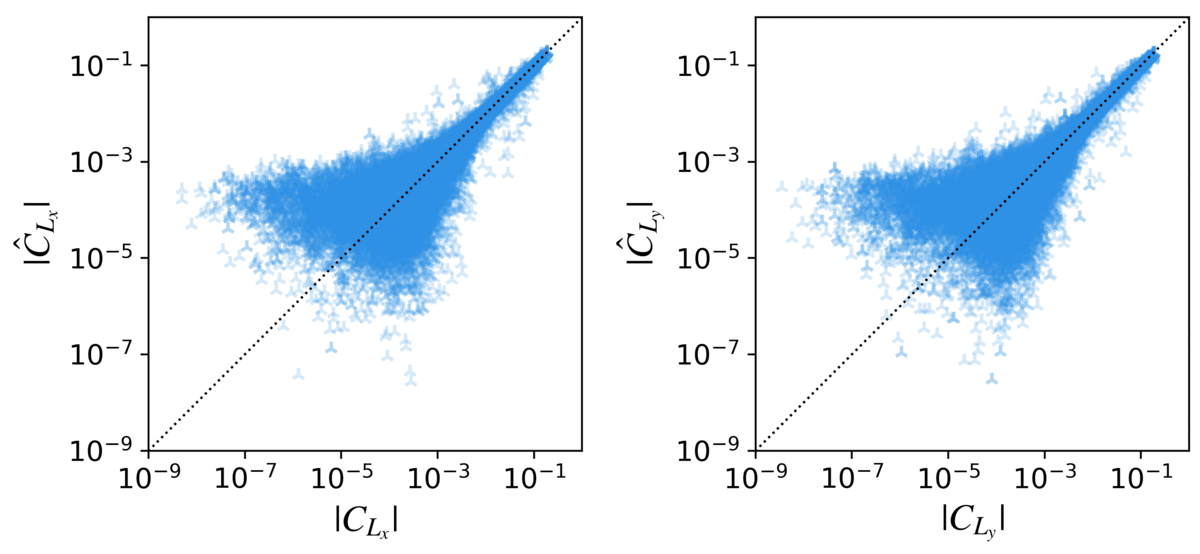}
        \label{f-linear}
    \end{subfigure}
    \begin{subfigure}{0.4\textwidth}
        \caption{}  
        \centering
        \includegraphics[alt={Two vertically stacked plots showing model error versus relative distance from the channel center. The top plot (blue) shows angular error in degrees, which remains near zero across most distances but exhibits high spikes and dispersion at the center (0) and mid-wall regions (0.5 to 0.6). The bottom plot (pink) shows relative magnitude error on a log scale, and exhibits a similar profile.}, width=\textwidth]{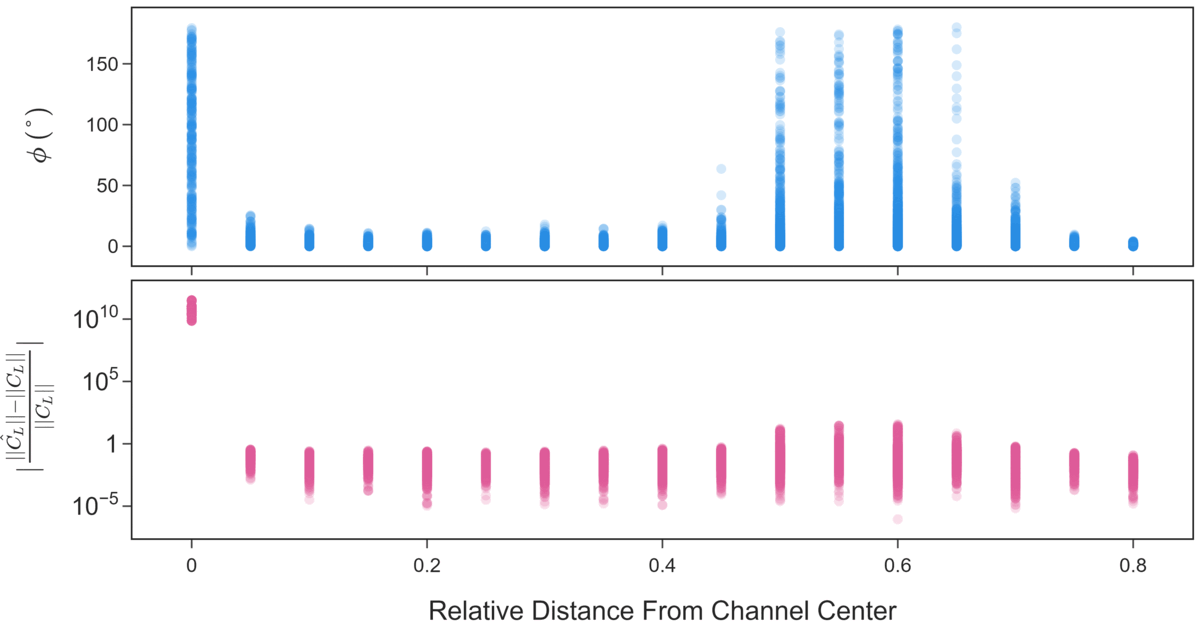}
        \label{f-error}
    \end{subfigure}
    \end{center}
    \vspace{-3mm}
    \caption[Model error distributions]{(a) Predicted lift coefficient vector components plotted against actual lift coefficient components. (b) Distribution of angle and relative magnitude errors within a square \qty{50}{\micro\meter} $\times$ \qty{50}{\micro\meter} microchannel. The relative distance from the channel centre is the relative distance of a particle along a line from the channel centre, through that particle, to the channel wall. This distance is zero at the centre of the channel and unity at any wall.}
    \label{f-error-plots}
    \end{figure}

    Plotting the predicted values of $C_{\texttt{L}_x}$ and $C_{\texttt{L}_y}$ against their actual values shows that the relative errors increase as the components approach zero (\cref{f-linear}). In other words, the smaller the $\mathbf{C}_\texttt{L}$ component, the harder it was for the model to predict. This led to a spatial correlation in the model errors: the largest errors occurred either close to the channel centre or at a specific intermediate distance between the centre and the walls (\cref{f-error}). These are both regions in which lift forces become vanishingly small, making them more difficult to predict. At the channel centre, vanishingly small forces are from the symmetry of the velocity profile. At the intermediate distance, they are due to a balance between shear gradient lift forces and wall-induced lift and it is at these lateral equilibrium positions where particles ultimately focus during inertial migration.\cite{DiCarlo2009b, Zhou2013}. The downstream simulations (\cref{f-simulation}) suggested that, practically, these errors were not large enough to prevent particles from stabilizing at their eventual equilibrium positions within the simulated IMDs.

    To avoid explicit dependence on geometric parameters, the non-dimensionalization step used in \cref{e-no-dim} differed slightly from literature and as a consequence our $C_{\texttt{L}_x}$ and $C_{\texttt{L}_y}$ values spanned more orders of magnitude (from \num{1e-10} to \num{1e-0}). Designing a loss function that is appropriate for a large range of extremely small target values centred around zero is challenging, as both absolute and relative errors have large drawbacks in such a setting.\cite{Botchkarev2019} To reduce the error on small $\mathbf{C}_\texttt{L}$ we tried several approaches: \begin{enumerate*}[label=(\alph*)]
        \item using relative loss functions, which led to gradient explosions and non-convergence;
        \item normalizing the target data, which lowered the upper limit of $C_{\texttt{L}_x}$ and $C_{\texttt{L}_y}$ values but did not affect the majority of the data close to zero;
        \item transforming the target data into logarithmic space, which introduced some challenges with negative and zero values, and resulted in worse models; and
        \item predicting angle and magnitude instead of vector components, which improved $\phi^{err}_{50}$ and $\lVert\mathbf{C}_\texttt{L}\rVert^{err}_{50}$ scores, but resulted in many more outliers and prediction artifacts, and thus much higher $\phi^{err}_{90}$ and $\lVert\mathbf{C}_\texttt{L}\rVert^{err}_{90}$ scores.
    \end{enumerate*}

    \section{Conclusions}
    \label{conclusions}
    IMDs offer low-cost, high-throughput alternative techniques for many traditional particle- and cell-separation tasks, but simulating them requires being able to predict particle migration, and thus particle lift forces, under a variety of geometric conditions. We exploited the underlying scaling laws governing particle migration to develop a novel model parametrization that has no explicit dependence on channel geometries. We demonstrated how FCNN models using our parametrization can generalize to geometries that were not seen in the training data. We further showed that these models can be readily incorporated into particle tracing simulations in software such as \texttt{COMSOL}, and result in particle trajectories which match physical experiments from literature. It is our hope that the model trained herein can be immediately useful to researchers running simulations, and that this alternative formulation for the lift force will open the door to further novel approaches as the AI and deep learning fields advance. 
    
\section*{Author Contributions}
    \textbf{Jesse Ward-Bond:} Conceptualization, Methodology, Investigation, Software, Visualization, Writing - Original Draft Creation. \textbf{Ali Mashhadian:} Conceptualization, Methodology, Data Curation, Simulation, Validation. \textbf{Timothy Chan:} Conceptualization, Supervision, Writing - Review \& Editing. \textbf{Edmond Young:} Conceptualization, Supervision, Writing - Review \& Editing. 

\section*{Conflicts of interest}
    There are no conflicts to declare.

\section*{Data availability}
    Code and model checkpoints are available at \href{https://github.com/jwardbond/MF-lift}{https://github.com/jwardbond/MF-lift}. Datasets and preprocessing steps are available on Zenodo at \href{https://doi.org/10.5281/zenodo.19377330}{10.5281/zenodo.19377330}.    

\section*{Acknowledgements}
    This research was supported by the NSERC Discovery Grant (RGPIN-2019-05885) held by Edmond Young and by the Ontario Graduate Scholarship, Queen Elizabeth II Graduate Scholarship in Science \& Technology, and Vector Scholarship in AI held by Jesse Ward-Bond.

\bibliographystyle{rsc} 
\bibliography{references} 

@article{DiCarlo2009b,
  title={{P}article segregation and dynamics in confined flows},
  author={Di Carlo, Dino and Edd, Jon F and Humphry, Katherine J and Stone, Howard A and Toner, Mehmet},
  journal={Phys. Rev. Lett.},
  volume={102},
  number={9},
  pages={094503},
  year={2009},
  publisher={APS}
}

@article{DiCarlo2007,
  title={{C}ontinuous inertial focusing, ordering, and separation of particles in microchannels},
  author={Di Carlo, Dino and Irimia, Daniel and Tompkins, Ronald G and Toner, Mehmet},
  journal={Proc. Natl. Acad. Sci. U. S. A.},
  volume={104},
  number={48},
  pages={18892--18897},
  year={2007},
  publisher={National Acad Sciences}, 
  doi={10.1073/pnas.0704958104}
}

@article{Hood2015, 
    title={{I}nertial migration of a rigid sphere in three-dimensional {P}oiseuille flow}, 
    volume={765}, 
    doi={10.1017/jfm.2014.739}, 
    journal={J. Fluid Mech.}, 
    publisher={Cambridge University Press}, 
    author={Hood, Kaitlyn and Lee, Sungyon and Roper, Marcus}, 
    year={2015}, 
    pages={452--479}
}

@article{Shamloo2018,
  title={{I}nertial particle focusing in serpentine channels on a centrifugal platform},
  author={Shamloo, Amir and Mashhadian, Ali},
  journal={Phys. Fluids},
  volume={30},
  number={1},
  pages={012002},
  year={2018},
  publisher={AIP Publishing LLC}
}

@article{Su2021,
  title={{M}achine learning assisted fast prediction of inertial lift in microchannels},
  author={Su, Jinghong and Chen, Xiaodong and Zhu, Yongzheng and Hu, Guoqing},
  journal={Lab Chip},
  volume={21},
  number={13},
  pages={2544--2556},
  year={2021},
  publisher={Royal Society of Chemistry}
}

@article{Segre1961,
  title={{R}adial particle displacements in {P}oiseuille flow of suspensions},
  author={Segr\'{e}, G and Silberberg, A},
  journal={Nature},
  volume={189},
  number={4760},
  pages={209--210},
  year={1961},
  publisher={Springer}
}

@article{Zhou2013,
  title={{F}undamentals of inertial focusing in microchannels},
  author={Zhou, Jian and Papautsky, Ian},
  journal={Lab Chip},
  volume={13},
  number={6},
  pages={1121--1132},
  year={2013},
  publisher={Royal Society of Chemistry}
}

@article{Moloudi2018,
  title={{I}nertial particle focusing dynamics in a trapezoidal straight microchannel: {A}pplication to particle filtration},
  author={Moloudi, Reza and Oh, Steve and Yang, Chun and Ebrahimi Warkiani, Majid and Naing, May Win},
  journal={Microfluid. Nanofluid.},
  volume={22},
  pages={1--14},
  year={2018},
  publisher={Springer}
}

@article{Liu2015,
  title={{I}nertial focusing of spherical particles in rectangular microchannels over a wide range of {R}eynolds numbers},
  author={Liu, Chao and Hu, Guoqing and Jiang, Xingyu and Sun, Jiashu},
  journal={Lab Chip},
  volume={15},
  number={4},
  pages={1168--1177},
  year={2015},
  publisher={Royal Society of Chemistry}
}

@InProceedings{sutskever2013,
  title = 	 {{O}n the importance of initialization and momentum in deep learning},
  author = 	 {Sutskever, Ilya and Martens, James and Dahl, George and Hinton, Geoffrey},
  booktitle = 	 {Proceedings of the 30th International Conference on Machine Learning},
  pages = 	 {1139--1147},
  year = 	 {2013},
  editor = 	 {Dasgupta, Sanjoy and McAllester, David},
  volume = 	 {28},
  number =       {3},
  series = 	 {Proceedings of Machine Learning Research},
  address = 	 {Atlanta, Georgia, USA},
  month = 	 {17--19 Jun},
  publisher =    {PMLR},
  url = 	 {https://proceedings.mlr.press/v28/sutskever13.html},
}

@article{Botchkarev2019,
	doi = {10.28945/4184},
	url = {https://doi.org/10.28945%2F4184},
	year = 2019,
	publisher = {Informing Science Institute},
	volume = {14},
	pages = {045--076},
	author = {Alexei Botchkarev},
	title = {{A} {N}ew {T}ypology {D}esign of {P}erformance {M}etrics to {M}easure {E}rrors in {M}achine {L}earning {R}egression {A}lgorithms},
	journal = {Interdisciplinary Journal of Information, Knowledge, and Management}
}

@article{Buckingham1914,
  title={{O}n physically similar systems; illustrations of the use of dimensional equations},
  author={Buckingham, Edgar},
  journal={Phys. Rev.},
  volume={4},
  number={4},
  pages={345},
  year={1914},
  publisher={APS}
}

@inproceedings{lundberg2017,
author = {Lundberg, Scott M. and Lee, Su-In},
title = {A unified approach to interpreting model predictions},
year = {2017},
isbn = {9781510860964},
publisher = {Curran Associates Inc.},
address = {Red Hook, NY, USA},
booktitle = {Proceedings of the 31st International Conference on Neural Information Processing Systems},
pages = {4768–4777},
numpages = {10},
location = {Long Beach, California, USA},
series = {NIPS'17}
}

@article{xiang2022,
  title={{I}nertial microfluidics: current status, challenges, and future opportunities},
  author={Xiang, Nan and Ni, Zhonghua},
  journal={Lab Chip},
  volume={22},
  number={24},
  pages={4792--4804},
  year={2022},
  publisher={Royal Society of Chemistry}, 
  doi={10.1039/D2LC00722C}
}

@article{Asmolov2018,
  title={{I}nertial focusing of finite-size particles in microchannels},
  author={Asmolov, Evgeny S and Dubov, Alexander L and Nizkaya, Tatiana V and Harting, Jens and Vinogradova, Olga I},
  journal={J. Fluid Mech.},
  volume={840},
  pages={613--630},
  year={2018},
  publisher={Cambridge University Press},
  doi={10.1017/jfm.2018.95}
}

@article{Su2023,
  title={{N}ew explicit formula for inertial lift in confined flows},
  author={Su, Jinghong and Zheng, Xu and Hu, Guoqing},
  journal={Phys. Fluids},
  volume={35},
  number={9},
  year={2023},
  publisher={AIP Publishing},
  doi={10.1063/5.0168147},
  pages={092010}
}

@misc{Godbole2023,
  author = {Varun Godbole and George E. Dahl and Justin Gilmer and Christopher J. Shallue and Zachary Nado},
  title = {{D}eep {L}earning {T}uning {P}laybook},
  url = {http://github.com/google-research/tuning_playbook},
  year = {2023},
  note = {Version 1.0}
}

@article{Zhang2014,
  title={{I}nertial particle separation by differential equilibrium positions in a symmetrical serpentine micro-channel},
  author={Zhang, Jun and Yan, Sheng and Sluyter, Ronald and Li, Weihua and Alici, Gursel and Nguyen, Nam-Trung},
  journal={Sci. Rep.},
  volume={4},
  number={1},
  pages={4527},
  year={2014},
  publisher={Nature Publishing Group UK London}
}

@article{Akhbari2024,
  title={{A} correlation-based approach for fast prediction of the equilibrium position of particles within a spiral microchannel},
  author={Akhbari, Mitra and Nasr Esfahany, Mohsen},
  journal={Indian Chem. Eng.},
  pages={1--17},
  year={2024},
  publisher={Taylor \& Francis}
}

@article{Bazaz2020,
  title={{C}omputational inertial microfluidics: {A} review},
  author={Bazaz, Sajad Razavi and Mashhadian, Ali and Ehsani, Abbas and Saha, Suvash Chandra and Kr{\"u}ger, Timm and Warkiani, Majid Ebrahimi},
  journal={Lab Chip},
  volume={20},
  number={6},
  pages={1023--1048},
  year={2020},
  publisher={Royal Society of Chemistry}
}

@article{Li2015,
  title={{D}ynamics of particle migration in channel flow of viscoelastic fluids},
  author={Li, Gaojin and McKinley, Gareth H and Ardekani, Arezoo M},
  journal={J. Fluid Mech.},
  volume={785},
  pages={486--505},
  year={2015},
  publisher={Cambridge University Press}
}

@article{Hur2012,
  title={{L}abel-free enrichment of adrenal cortical progenitor cells using inertial microfluidics},
  author={Hur, Soojung Claire and Brinckerhoff, Tatiana Z and Walthers, Christopher M and Dunn, James CY and Di Carlo, Dino},
  year={2012},
  journal={PLOS ONE},
  pages={46550},
  publisher={Public Library of Science San Francisco, USA}
}

@article{Yoon2009,
  title={{S}ize-selective separation of micro beads by utilizing secondary flow in a curved rectangular microchannel},
  author={Yoon, Dong Hyun and Ha, Jin Bong and Bahk, Yoen Kyung and Arakawa, Takahiro and Shoji, Shuichi and Go, Jeung Sang},
  journal={Lab Chip},
  volume={9},
  number={1},
  pages={87--90},
  year={2009},
  publisher={Royal Society of Chemistry}
}

@article{Guan2013,
  title={{S}piral microchannel with rectangular and trapezoidal cross-sections for size based particle separation},
  author={Guan, Guofeng and Wu, Lidan and Bhagat, Ali Asgar and Li, Zirui and Chen, Peter CY and Chao, Shuzhe and Ong, Chong Jin and Han, Jongyoon},
  journal={Sci. Rep.},
  volume={3},
  number={1},
  pages={1475},
  year={2013},
  publisher={Nature Publishing Group UK London}
}

@article{Park2009,
  title={{C}ontinuous focusing of microparticles using inertial lift force and vorticity via multi-orifice microfluidic channels},
  author={Park, Jae-Sung and Song, Suk-Heung and Jung, Hyo-Il},
  journal={Lab Chip},
  volume={9},
  number={7},
  pages={939--948},
  year={2009},
  publisher={Royal Society of Chemistry}
}

@article{Mcintyre2022,
  title={{M}achine learning for microfluidic design and control},
  author={McIntyre, David and Lashkaripour, Ali and Fordyce, Polly and Densmore, Douglas},
  journal={Lab Chip},
  volume={22},
  number={16},
  pages={2925--2937},
  year={2022},
  publisher={Royal Society of Chemistry}
}

@article{Owen2024,
  title={{A}ccelerating the development of inertial microfluidic devices using numerical modelling and machine learning},
  author={Owen, Benjamin},
  journal={Front. Lab Chip Technol.},
  volume={3},
  pages={1328004},
  year={2024},
  publisher={Frontiers Media SA}
}

@article{Safari2024,
  title={{S}piral{D}esigner: {A}n {A}{I}-assisted design interface for efficient separation of neutrally buoyant and non-buoyant particles using spiral microfluidic devices},
  author={Safari, Morteza and Abbasi, Pezhman and Momeni, Seyedali and Farahani, Mahdieh Shahrabi and Safari, Hanieh},
  journal={Chem. Eng. Sci.},
  volume={297},
  pages={120301},
  year={2024},
  publisher={Elsevier}
}

@article{Lee2011,
  title={{I}nertial separation in a contraction--expansion array microchannel},
  author={Lee, Myung Gwon and Choi, Sungyoung and Park, Je-Kyun},
  journal={J. Chromatogr. A},
  volume={1218},
  number={27},
  pages={4138--4143},
  year={2011},
  publisher={Elsevier}
}

@article{Nesterov1983,
  title={{A} method of solving a convex programming problem with convergence rate $O$(1/k2)},
  author={Nesterov, Y},
  journal={Proc. USSR Acad. Sci.},
  year={1983},
  volume={269},
  pages={3}
}

@misc{Bai2019,
    author = {Bai, Junjie and Lu, Fang and Zhang, Ke and others},
    title = {{O}{N}{N}{X}: {O}pen {N}eural {N}etwork {E}xchange},
    year = {2019},
    publisher = {GitHub},
    journal = {GitHub repository},
    howpublished = {\url{https://github.com/onnx/onnx}},
    commit = {94d238d96e3fb3a7ba34f03c284b9ad3516163be}
}

@article{Zhou2018,
  title={{S}heathless inertial cell focusing and sorting with serial reverse wavy channel structures},
  author={Zhou, Yinning and Ma, Zhichao and Ai, Ye},
  journal={Microsyst. Nanoeng.},
  volume={4},
  number={1},
  pages={5},
  year={2018},
  publisher={Nature Publishing Group UK London}
}

@article{Dai2015,
  title={{E}uler--{R}odrigues formula variations, quaternion conjugation and intrinsic connections},
  author={Dai, Jian S},
  journal={Mech. Mach. Theory},
  volume={92},
  pages={144--152},
  year={2015},
  publisher={Elsevier}
}

@article{Wu2016,
  title={{C}ontinuous inertial microparticle and blood cell separation in straight channels with local microstructures},
  author={Wu, Zhenlong and Chen, Yu and Wang, Moran and Chung, Aram J},
  journal={Lab Chip},
  volume={16},
  number={3},
  pages={532--542},
  year={2016},
  publisher={Royal Society of Chemistry}
}

@article{Kim2022,
  title={{P}article focusing in a straight microchannel with non-rectangular cross-section},
  author={Kim, Uihwan and Kwon, Joo-Yong and Kim, Taehoon and Cho, Younghak},
  journal={Micromachines},
  volume={13},
  number={2},
  pages={151},
  year={2022},
  publisher={MDPI}
}

@article{Warkiani2016,
  title={{U}ltra-fast, label-free isolation of circulating tumor cells from blood using spiral microfluidics},
  author={Warkiani, Majid Ebrahimi and Khoo, Bee Luan and Wu, Lidan and Tay, Andy Kah Ping and Bhagat, Ali Asgar S and Han, Jongyoon and Lim, Chwee Teck},
  journal={Nat. Protoc.},
  volume={11},
  number={1},
  pages={134--148},
  year={2016},
  publisher={Nature Publishing Group UK London}
}

@article{Nivedita2017,
  title={{D}ean flow dynamics in low-aspect ratio spiral microchannels},
  author={Nivedita, Nivedita and Ligrani, Phillip and Papautsky, Ian},
  journal={Sci. Rep.},
  volume={7},
  number={1},
  pages={44072},
  year={2017},
  publisher={Nature Publishing Group UK London}
}

@article{zhang2016,
  title={{F}undamentals and applications of inertial microfluidics: a review},
  author={Zhang, Jun and Yan, Sheng and Yuan, Dan and Alici, Gursel and Nguyen, Nam-Trung and Warkiani, Majid Ebrahimi and Li, Weihua},
  journal={Lab Chip},
  volume={16},
  number={1},
  pages={10--34},
  year={2016},
  publisher={Royal Society of Chemistry}
}

@article{jiang2022,
  title={{I}ntegrated microfluidic handheld cell sorter for high-throughput label-free malignant tumor cell sorting},
  author={Jiang, Fengtao and Xiang, Nan},
  journal={Anal. Chem.},
  volume={94},
  number={3},
  pages={1859--1866},
  year={2022},
  publisher={ACS Publications}
}

@article{jeon2022,
  title={{E}ngineering a deformation-free plastic spiral inertial microfluidic system for {C}{H}{O} cell clarification in biomanufacturing},
  author={Jeon, Hyungkook and Kwon, Taehong and Yoon, Junghyo and Han, Jongyoon},
  journal={Lab Chip},
  volume={22},
  number={2},
  pages={272--285},
  year={2022},
  publisher={Royal Society of Chemistry}
}

@article{volpe2019,
  title={{P}olymeric fully inertial lab-on-a-chip with enhanced-throughput sorting capabilities},
  author={Volpe, Annalisa and Pai{\`e}, Petra and Ancona, Antonio and Osellame, Roberto},
  journal={Microfluid. Nanofluid.},
  volume={23},
  number={3},
  pages={37},
  year={2019},
  publisher={Springer}
}

@article{guckenberger2015,
  title={{M}icromilling: a method for ultra-rapid prototyping of plastic microfluidic devices},
  author={Guckenberger, David J and De Groot, Theodorus E and Wan, Alwin MD and Beebe, David J and Young, Edmond WK},
  journal={Lab Chip},
  volume={15},
  number={11},
  pages={2364--2378},
  year={2015},
  publisher={Royal Society of Chemistry}
}

@article{sia2003,
  title={{M}icrofluidic devices fabricated in poly (dimethylsiloxane) for biological studies},
  author={Sia, Samuel K and Whitesides, George M},
  journal={Electrophoresis},
  volume={24},
  number={21},
  pages={3563--3576},
  year={2003},
  publisher={Wiley Online Library}
}

@article{hwang2019,
  title={{M}icrochannel fabrication on glass materials for microfluidic devices},
  author={Hwang, Jihong and Cho, Young Hak and Park, Min Soo and Kim, Bo Hyun},
  journal={International Journal of Precision Engineering and Manufacturing},
  volume={20},
  number={3},
  pages={479--495},
  year={2019},
  publisher={Springer}
}

@article{grimes2008,
  title={{S}hrinky-{D}ink microfluidics: rapid generation of deep and rounded patterns},
  author={Grimes, Anthony and Breslauer, David N and Long, Maureen and Pegan, Jonathan and Lee, Luke P and Khine, Michelle},
  journal={Lab Chip},
  volume={8},
  number={1},
  pages={170--172},
  year={2008},
  publisher={Royal Society of Chemistry}
}

@article{tang2020,
  title={{C}hannel innovations for inertial microfluidics},
  author={Tang, Wenlai and Zhu, Shu and Jiang, Di and Zhu, Liya and Yang, Jiquan and Xiang, Nan},
  journal={Lab Chip},
  volume={20},
  number={19},
  pages={3485--3502},
  year={2020},
  publisher={Royal Society of Chemistry}
}

@misc{Hamdi2022,
      title={microAI: A machine learning tool for fast calculation of lift coefficients in microchannels}, 
      author={Erfan Hamdi and Rasool Dezhkam and Amir Shamloo and Ali Mashhadian},
      year={2022},
      eprint={2210.11591},
      archivePrefix={arXiv},
      primaryClass={physics.flu-dyn},
      url={https://arxiv.org/abs/2210.11591}, 
}

\clearpage
\begin{appendix}
\section*{Supplementary Information}
\label{s-supplemental}
\setcounter{subsection}{0}
\renewcommand{\thesubsection}{S\arabic{subsection}}
\renewcommand{\theequation}{S\arabic{equation}}
\renewcommand{\thefigure}{S\arabic{figure}}
\setcounter{equation}{0}
\setcounter{figure}{0}

    \subsection{Taylor Series Approximation}
    \label{s-taylor}

    \Cref{f-taylor} shows that a second-order Taylor series expansion can almost completely recover the velocity profile around the circumference of an ``imaginary'' (non-interacting) spherical particle within a straight rectangular channel.

    \begin{figure}[h]
    \centering
        \includegraphics[width=0.48\textwidth]{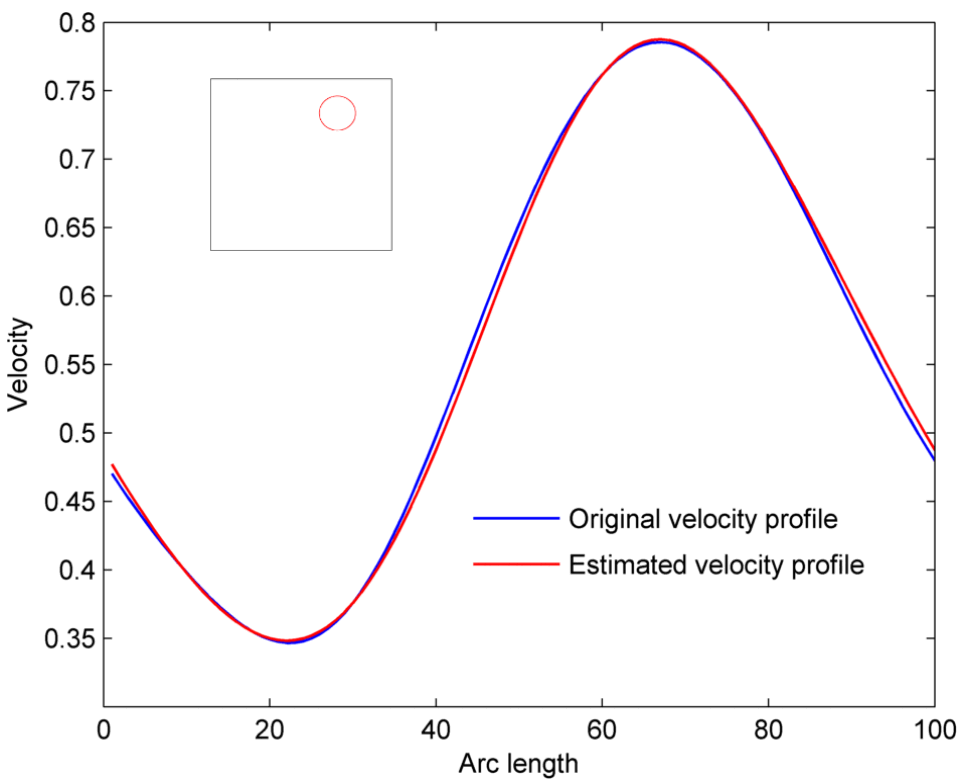}
    \caption[Second order Taylor expansion around a particle in a rectangular channel]{The second-order Taylor series expansion of the local velocity profile around the circumference of a particle in a rectangular channel.}
    \label{f-taylor}
    \end{figure}

    \subsection{$\mathbf{C}_\texttt{L}$ Conversion}
    \label{s-conversion}
    To convert from the $\mathbf{C}_\texttt{L}$ formulation used in \citet{Su2021} to our current methodology, we use

    \begin{equation}
        \mathbf{C}_\texttt{L}  = \mathbf{C}_\texttt{L}^{Su}\cdot \frac{a^2 U_m^2}{w^2 H^2},
    \end{equation}

    \noindent where $U_m$ is the maximum (centerline) velocity in the channel.

    \subsection{Generalization to Arbitrary 3D Orientations}
    \label{s-rotation}
    Note that in the following, we use $(x,y,z)$ to represent the \textit{global} (lab referenced) coordinate system.
    
    As established in \cref{e-no-dim2}, the model formulation assumes that velocity vectors in the undisturbed flow are strictly aligned with the $z$-direction, rendering all $z$-gradients zero. While this performs well for fully developed flows in straight channels, it is insufficient for particle tracing in channels with arbitrary orientations (e.g., curved or serpentine channels) where the local fluid velocity vector continuously changes direction. To address this, we introduce a rotational mapping process to generalize the model for any 3D orientation using Rodrigues' rotation formula.\cite{Dai2015}. This mapping aligns the local fluid velocity vector $\mathbf{u} = \{u, v, w\}^\top$ with the global $z$-direction (represented by the unit vector $\mathbf{\hat{n}}_z = \{0, 0, 1\}^\top$). Once rotated, we can calculate the $\mathbf{C}'_{\texttt{L}_{XY}}$---the lift force coefficient in the rotated reference frame---and then perform the inverse rotation to map this back to $\mathbf{C}_{\texttt{L}_{XYZ}}$, the 3D lift force vector in the original reference frame.

    We begin by constructing the rotation matrix, $\mathbf{R}_M$. The normalized velocity vector is defined as 
    \begin{equation}
        \mathbf{\hat{n}} = \frac{\mathbf{u}}{\|\mathbf{u}\|} = \left( \frac{u}{\|\mathbf{u}\|}, \frac{v}{\|\mathbf{u}\|}, \frac{w}{\|\mathbf{u}\|} \right).
    \end{equation}
    The axis of rotation is therefore $\mathbf{\hat{n}} \times \mathbf{\hat{n}}_z$, and the angle of rotation is $\mathbf{\hat{n}} \cdot \mathbf{\hat{n}}_z$. We then formulate the skew-symmetric cross-product matrix $\mathbf{S}_C$ as:
    
    \begin{equation}
        \mathbf{S}_C = \begin{bmatrix} 0 & 0 & -\frac{u}{\|\mathbf{u}\|} \\ 0 & 0 & -\frac{v}{\|\mathbf{u}\|} \\ \frac{u}{\|\mathbf{u}\|} & \frac{v}{\|\mathbf{u}\|} & 0 \end{bmatrix}
    \end{equation}
    
    \noindent The rotation matrix $\mathbf{R}_M$ is then defined as:
    
    \begin{equation}
    \begin{split}
        \mathbf{R}_M &= \mathbf{I} + \mathbf{S}_C + \left( \frac{1 - \mathbf{\hat{n}} \cdot \mathbf{\hat{n}}_z}{\|\mathbf{\hat{n}} \times \mathbf{\hat{n}}_z\|^2} \right) \mathbf{S}_C^2 \\
        &= \mathbf{I} + \mathbf{S}_C + \left( \frac{1 - \frac{w}{\|\mathbf{u}\|}}{\left(\frac{u}{\|\mathbf{u}\|}\right)^2 + \left(\frac{v}{\|\mathbf{u}\|}\right)^2} \right) \mathbf{S}_C^2
    \end{split}
    \end{equation}
    
    \noindent where $\mathbf{I}$ is the $3 \times 3$ identity matrix. 
    
    Let $U = \|\mathbf{u}\|$ denote the magnitude of the velocity vector. The first-order gradient vector $\mathbf{G}_1$ and second-order gradient tensor $\mathbf{G}_2$ of the velocity field (in the global coordinate system) are defined as:
    
    \begin{equation}
        \mathbf{G}_1 = \begin{bmatrix} U_x \\ U_y \\ U_z \end{bmatrix}, \quad
        \mathbf{G}_2 = \begin{bmatrix} U_{xx} & U_{xy} & U_{xz} \\ U_{yx} & U_{yy} & U_{yz} \\ U_{zx} & U_{zy} & U_{zz} \end{bmatrix}
    \end{equation}
    
    \noindent Applying $\mathbf{R}_M$, we map these gradients into the $xy$-plane. We use $\mathbf{u}' = \left\{u', v', w'\right\}$ to represent the rotated velocity vector. This vector is in the direction of $\mathbf{\hat{n}_z}$, and thus $\mathbf{u}' = \left\{0,0,w'\right\}$. The mapped gradients are:
    
    \begin{gather}
        \mathbf{G}'_1 = \mathbf{R}_M \mathbf{G}_1 = \begin{bmatrix} w'_{x} \\ w'_{y} \\ 0 \end{bmatrix} \\
        \mathbf{G}'_2 = \mathbf{R}_M \mathbf{G}_2 \mathbf{R}_M^\top = \begin{bmatrix} w'_{xx} & w'_{xy} & 0 \\ w'_{yx} & w'_{yy} & 0 \\ 0 & 0 & 0 \end{bmatrix}
    \end{gather}
    
    \noindent Because the gradients of velocity along the streamline are negligible compared to the cross-sectional gradients ($X, Y$), all terms containing $z$ derivatives are neglected. We can therefore reduce the mapped gradients to the 2D planar features required by the model:
    
    \begin{equation}
        \mathbf{G}'_1 = \begin{bmatrix} w'_{x} \\ w'_{y} \end{bmatrix}, \quad
        \mathbf{G}'_2 = \begin{bmatrix} w'_{xx} & w'_{xy} \\ w'_{yx} & w'_{yy} \end{bmatrix}
    \end{equation}
    
    \noindent These spatial derivatives serve as inputs to the neural network. The planar lift force coefficient is predicted using:
    
    \begin{equation}
        \mathbf{C}'_{\texttt{L}_{XY}} = f\left( Re'_p, \frac{w'_{x}a}{w'}, \frac{w'_{y}a}{w'}, \frac{w'_{xx}a^2}{w'}, \frac{w'_{yy}a^2}{w'}, \frac{w'_{xy}a^2}{w'} \right)
    \end{equation}
    
    \noindent where $Re_p = \frac{\rho w' a}{\mu}$, Finally, the resultant 3D lift force coefficient is recovered by applying the inverse rotation matrix $\mathbf{R}_M^\top$, and the 3D lift force is calculated as:
    
    \begin{align}
        \mathbf{C}_{\texttt{L}_{XYZ}} &= \mathbf{R}_M^\top \mathbf{C}'_{\texttt{L}_{XY}} \\
        \mathbf{F}_{\texttt{L}_{XYZ}} &= \rho U^2 a^2 \mathbf{C}_{\texttt{L}_{XYZ}}
    \end{align}

    \clearpage
    \subsection{Principal Component Analysis}
    Plotting the first two principal components of the training data shows that the triangular and trapezoidal datasets have wider distributions than the rectangular and semicircular datasets (\cref{f-pca}).

    \begin{figure}[h]
    \centering
        \includegraphics[width=0.8\textwidth]{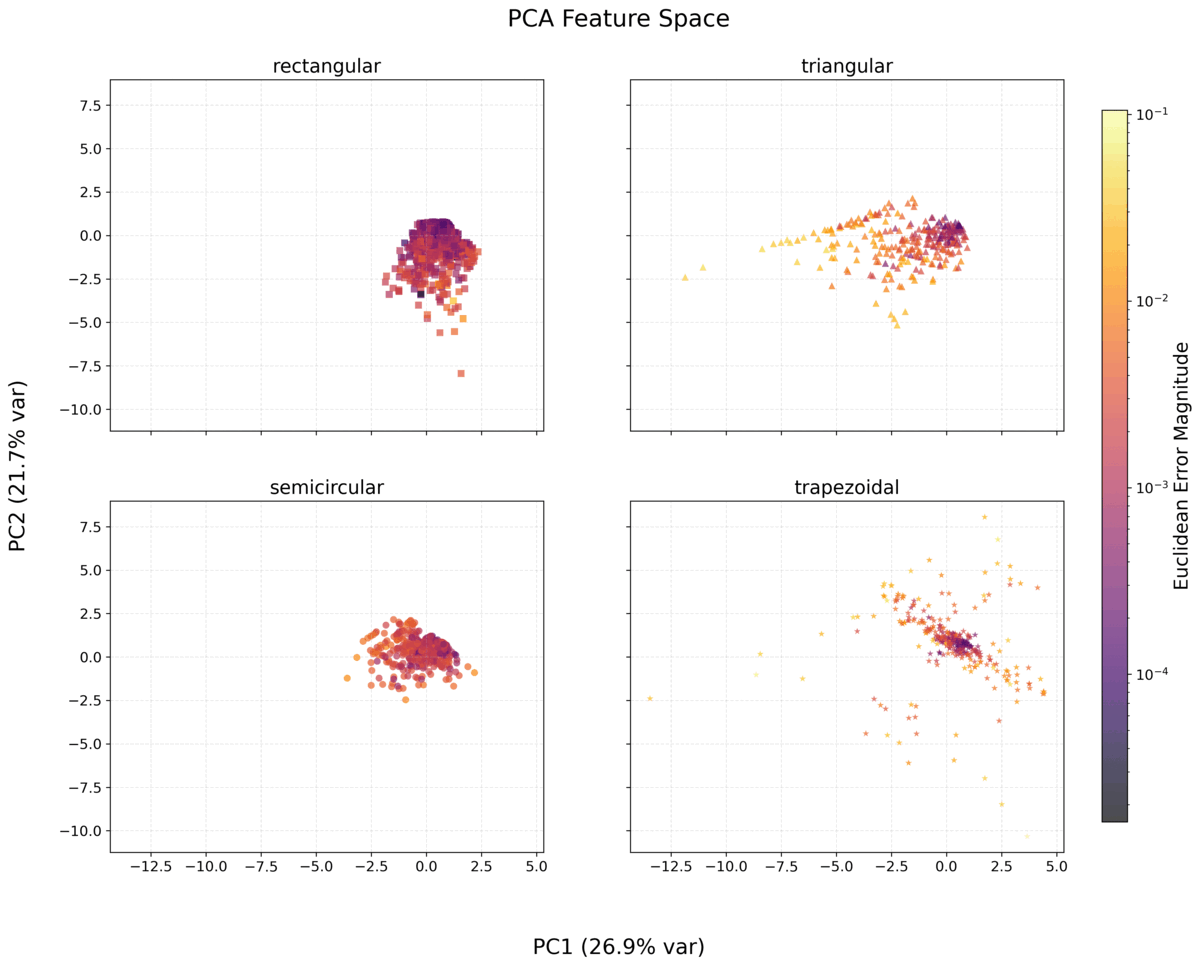}
    \caption[Pairwise adversarial validation]{The first two principal components of the training data, separated by geometry type and coloured by prediction error.}
    \label{f-pca}
    \end{figure}

    \clearpage
    \subsection{Numerical generalization results}

    \renewcommand{\arraystretch}{1.25}
    \begin{table*}[h]
    \centering\small
    \caption[Generalization of the best trained model to other geometries]{Generalization of trained model to other geometries. Values are reported as Median (90th Percentile). Models were retrained completely when additional geometries were added to the training set. $\mathcal{R, T, S, P}$, represent the rectangular, triangular, semicircular, and trapezoidal datasets, respectively. Superscripts $tr$ and $ts$ represent training and testing subsets.}
    \label{t-generalization}
    \begin{tabular*}{\textwidth}{@{\extracolsep{\fill}}clllll}
    \hline
    \multirow{2}{*}[-1mm]{\textbf{Metric}} & \multirow{2}{*}[-1mm]{\textbf{Training Set}} & \multicolumn{4}{c}{\textbf{Test Set}} \\
    \cmidrule(lr){3-6}
    & & \multicolumn{1}{c}{$\mathcal{R}^{ts}$} & \multicolumn{1}{c}{$\mathcal{T}^{ts}$} & \multicolumn{1}{c}{$\mathcal{S}^{ts}$} & \multicolumn{1}{c}{$\mathcal{P}^{ts}$} \\
    \hline
    \multirow{4}{*}{\textbf{MSE}}           
        & $\{\mathcal{R}\}^{tr}$       & \num{2.27e-6} & \num{8.83e-5} & \num{5.18e-6} & \num{7.44e-5} \\
        & $\{\mathcal{R,T}\}^{tr}$     & \num{2.38e-6} & \num{1.53e-6} & \num{2.85e-6} & \num{8.22e-5} \\
        & $\{\mathcal{R,T,S}\}^{tr}$   & \num{2.57e-6} & \num{1.75e-6} & \num{2.02e-6} & \num{1.03e-4} \\
        & $\{\mathcal{R,T,S,P}\}^{tr}$ & \num{3.58e-6} & \num{3.94e-6} & \num{2.68e-6} & \num{3.89e-5} \\
        &                           &                   &                   &                   & \\          
        
    \multirow{4}{*}{\thead{$\phi^{err}_{50} \ (\phi^{err}_{90})$ \\ (\unit{\degree})}}
        & $\{\mathcal{R}\}^{tr}$       & 4.6 (20.7)  & 7.6 (51.0)  & 12.8 (47.1) & 5.9 (58.8)  \\
        & $\{\mathcal{R,T}\}^{tr}$     & 4.7 (22.1)  & 3.2 (13.5)  & 12.3 (47.4) & 6.2 (56.7)  \\
        & $\{\mathcal{R,T,S}\}^{tr}$   & 5.4 (22.3)  & 3.3 (14.4)  & 9.7 (39.3)  & 6.9 (50.8)  \\
        & $\{\mathcal{R,T,S,P}\}^{tr}$ & 7.0 (29.1)  & 4.8 (20.3)  & 11.6 (44.8) & 6.2 (46.5)  \\
        &                           &                   &                   &                   & \\
                                        
    \multirow{4}{*}{\thead{$\lVert\mathbf{C}_\texttt{L}\rVert^{err}_{50} \ (\lVert\mathbf{C}_\texttt{L}\rVert^{err}_{90}) $ \\ (\%)}} 
        & $\{\mathcal{R}\}^{tr}$       & 8.7 (38.3)  & 21.2 (106.0) & 20.3 (65.6) & 13.0 (85.4) \\
        & $\{\mathcal{R,T}\}^{tr}$     & 9.5 (39.9)  & 6.6 (24.8)   & 19.0 (62.8) & 13.7 (81.1) \\
        & $\{\mathcal{R,T,S}\}^{tr}$   & 10.4 (51.5) & 7.0 (27.4)   & 16.2 (53.8) & 15.9 (103.3)\\ 
        & $\{\mathcal{R,T,S,P}\}^{tr}$ & 14.9 (71.2) & 11.0 (42.3)  & 18.1 (63.3) & 12.4 (81.4) \\
        &                           &                   &                   &                   & \\
        
    \hline
    \end{tabular*}
    \end{table*}

    \clearpage
    \subsection{Adversarial validation}
    \Cref{f-adversarial} shows the results of adversarial validation between all pairs of datasets. For each pair of datasets, we combined the two datasets and trained a histogram-based gradient boosting classification tree to predict which dataset each point belonged to using only the input features from \cref{e-no-dim2}. An area under the curve (AUC) score of \num{0.5} indicates that the model is unable to distinguish between the two datasets. Similarly, an AUC score of \num{1.0} shows that the model can perfectly separate the two datasets. We performed adversarial validation between the \textit{augmented} (rotated and flipped) datasets. 

    \begin{figure}[h]
    \centering
        \includegraphics[width=0.8\textwidth]{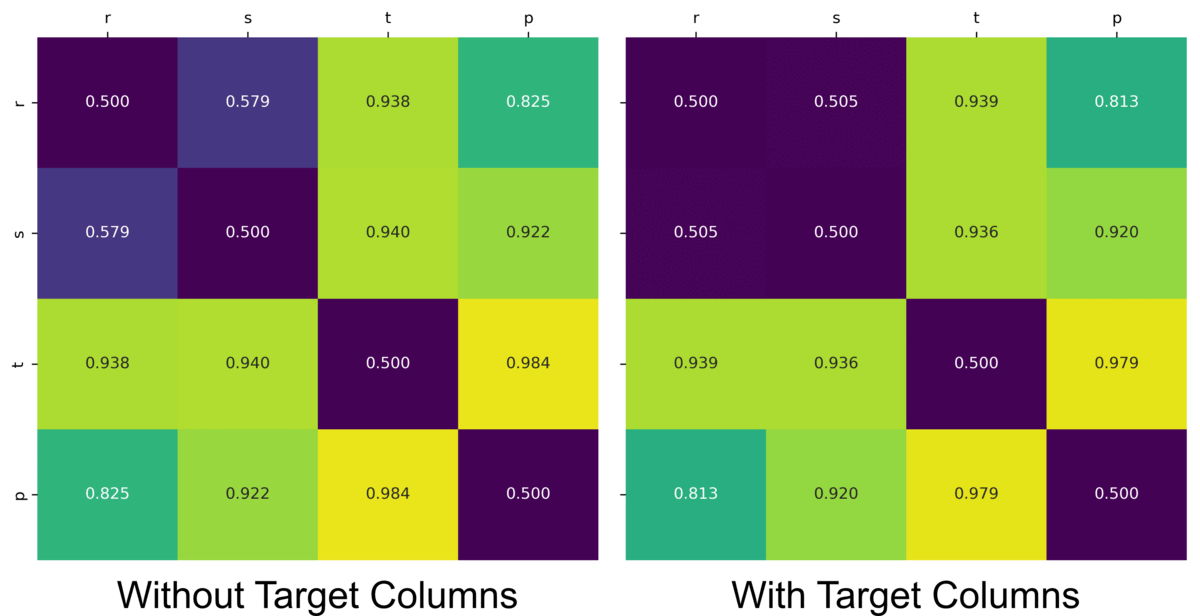}
    \caption[Pairwise adversarial validation]{Area under the curve (AUC) scores during adversarial validation of all input datasets using a histogram-based gradient boosting classification tree.}
    \label{f-adversarial}
    \end{figure}

    \clearpage
    \subsection{Additional Simulation Results}
    \label{s-supp-sim}
    
    \begin{figure}[h]
    \centering
        \includegraphics[width=0.5\textwidth]{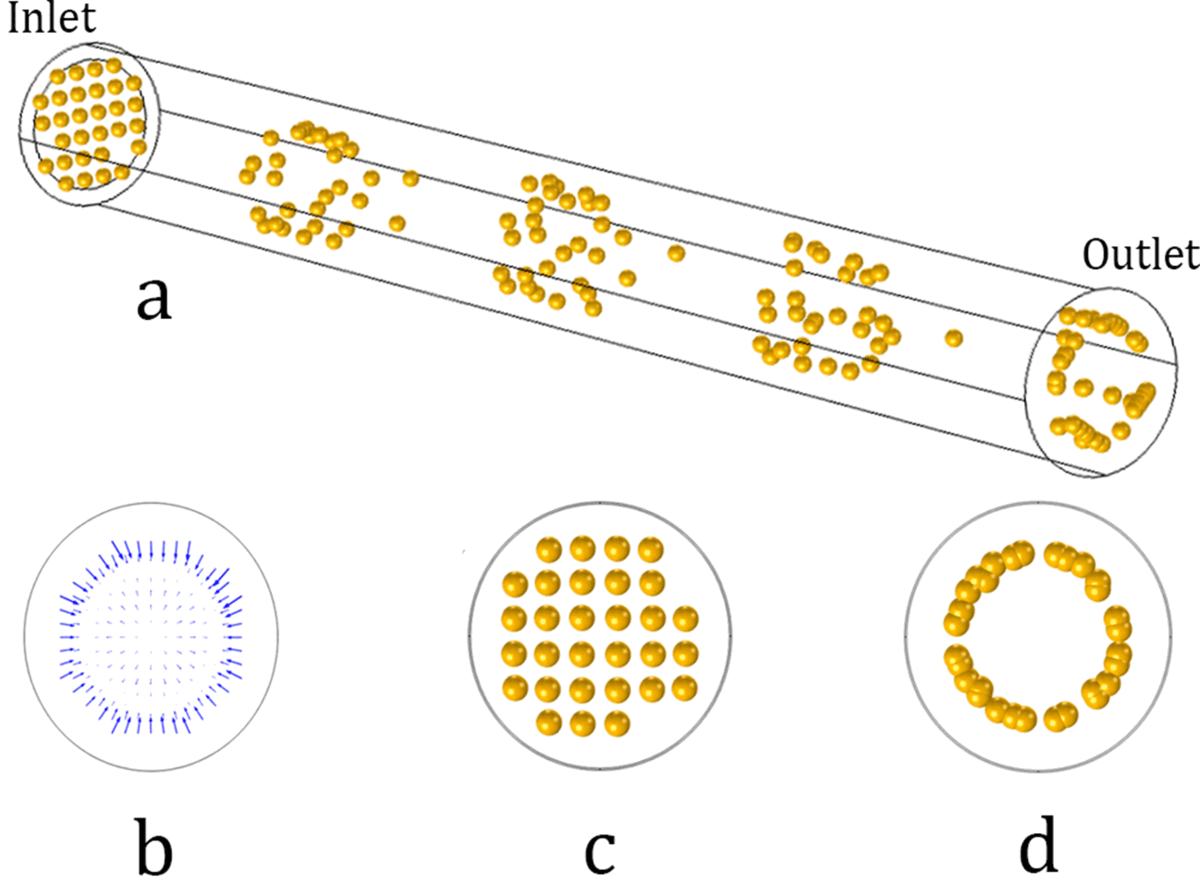}
    \caption[Semicircular channel simulations]{Inertial particle focusing in a straight circular microchannel using our geometry-free lift force model. The model was trained on straight channels with rectangular, triangular, semicircular, and trapezoidal cross-sections. The average fluid velocity in the channel is \qty{0.2}{\meter\per\second}, the particle size is \qty{10}{\micro\meter}, and the cross-section diameter is \qty{100}{\micro\meter}. a) Three-dimensional view of particles movement in the channel. b) Vector field plot of the predicted inertial lift forces. c) Particle positions at the channel inlet. d) Particle positions at the channel outlet.}
    \end{figure}

    \begin{figure}[h]
    \centering
        \includegraphics[width=0.5\textwidth]{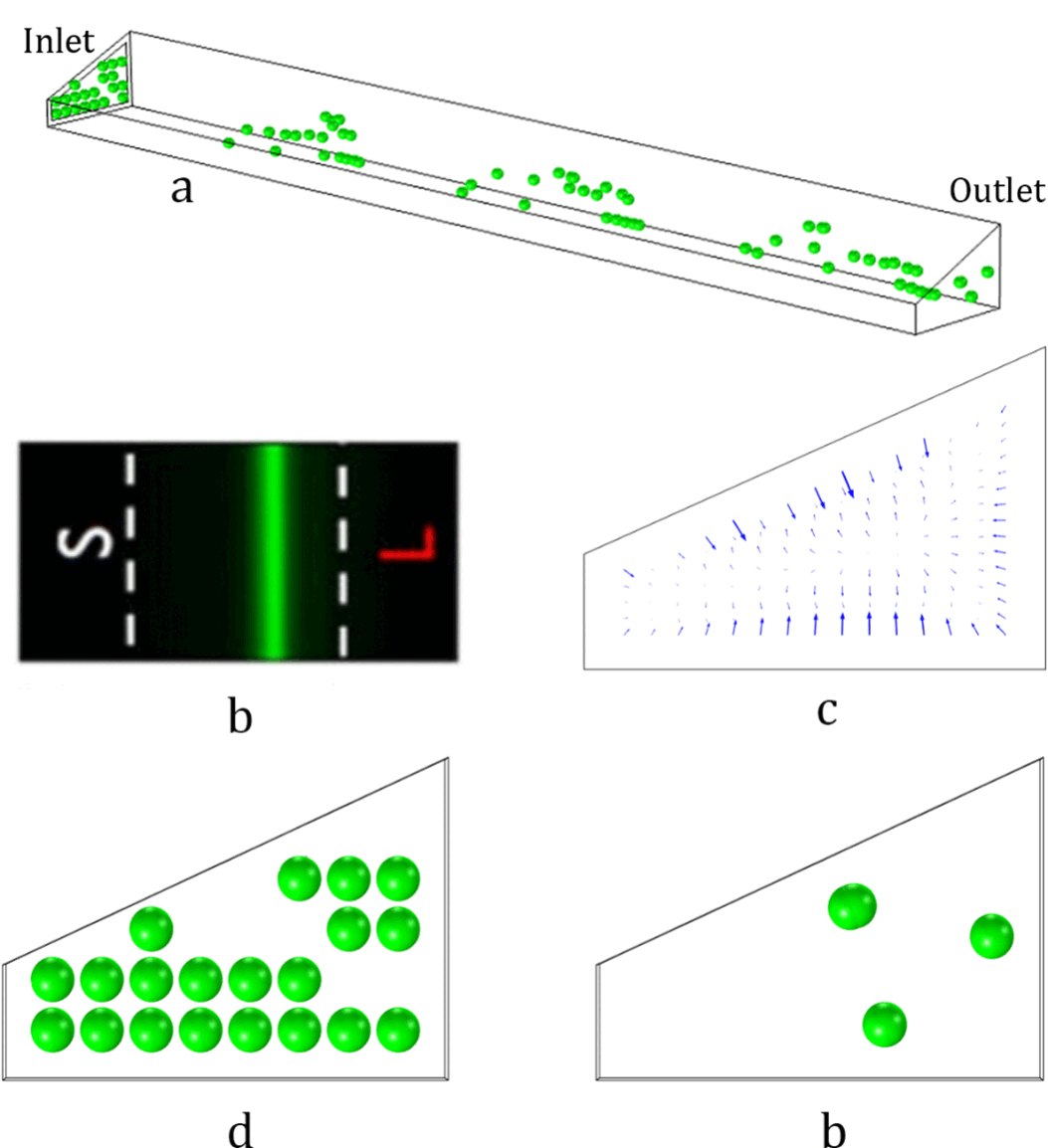}
    \caption[Semicircular channel simulations]{Inertial particle focusing in a straight trapezoidal microchannel using our geometry-free lift force model. The model was trained on straight channels with rectangular, triangular, semicircular, and trapezoidal cross-sections. The flow rate in the channel is \qty{0.2}{\milli\liter\per\minute}, the particle size is \qty{20}{\micro\meter}, and base, short height and long height of the trapezoid are \qty{200}{\micro\meter}, \qty{50}{\micro\meter}, and \qty{140}{\micro\meter}, respectively. b) Experimental results from literature for a straight trapezoidal channel under similar conditions. The green fluorescent lines represent the trajectories of the particles, and the white dashed lines delineate the channel walls. ``S'' and ``L'' indicate the short and long walls of the trapezoidal cross section, respectively. Adapted from \citet{Moloudi2018} with permission from Springer Nature. c) Vector field plot of the predicted inertial lift forces. d) Particle positions at the channel inlet. e) Particle positions at the channel outlet.}
    \end{figure}

    \begin{figure}[h]
    \centering
        \includegraphics[width=0.5\textwidth]{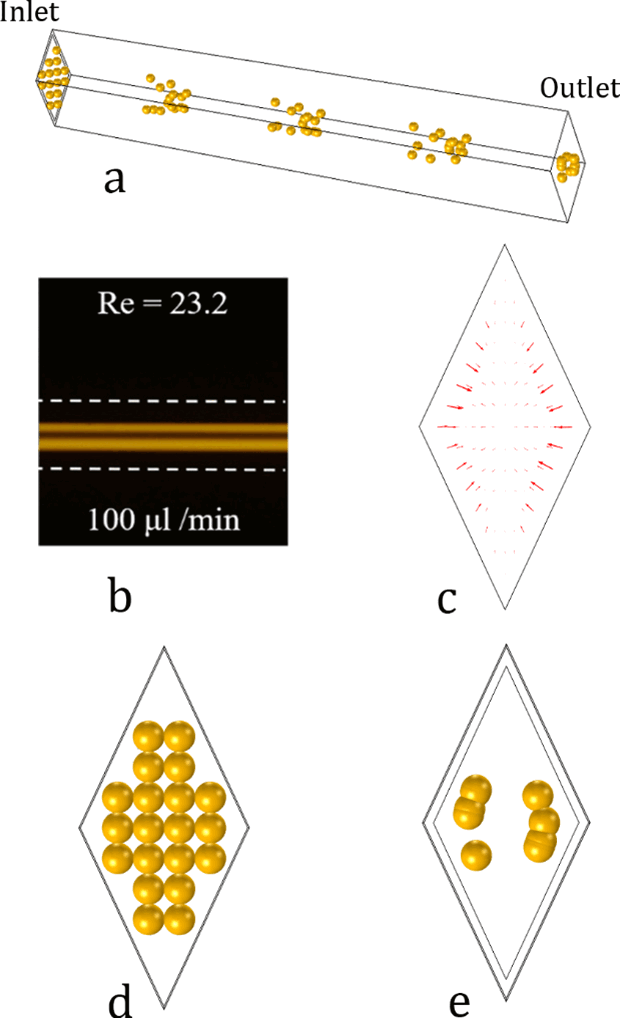}
    \caption[Semicircular channel simulations]{Inertial particle focusing in a straight rhombus microchannel using our geometry-free lift force model. The model was trained on straight channels with rectangular, triangular, semicircular, and trapezoidal cross-sections. The flow rate in the channel is \qty{0.1}{\milli\liter\per\minute}, and the particle size is \qty{13}{\micro\meter}. The diagonals of the rhombus measure \qty{68.4}{\micro\meter} and \qty{146.2}{\micro\meter}, respectively. b) Experimental results from literature for a straight rhombus channel under similar conditions. Adapted from "Particle Focusing in a Straight Microchannel with Non-Rectangular Cross-Section" by \citet{Kim2022} under CC BY 4.0 (\url{https://creativecommons.org/licenses/by/4.0/}). Image is a cropped version of the original. c) Vector field plot of the predicted inertial lift forces. d) Particle positions at the channel inlet. e) Particle positions at the channel outlet.}
    \end{figure}

    \begin{figure}[h]
    \centering
        \includegraphics[width=0.8\textwidth]{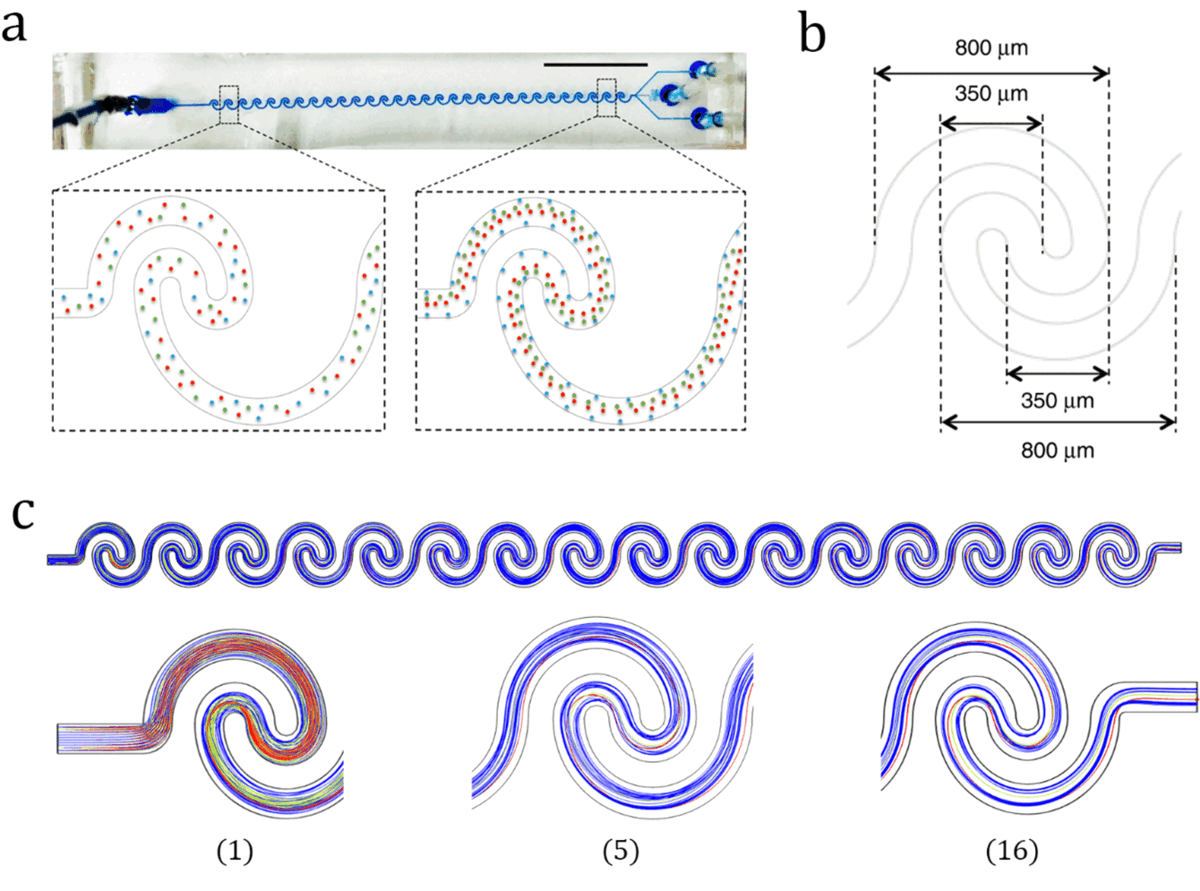}
    \caption[Semicircular channel simulations]{ Inertial separation of \qty{3}{\micro\meter} (blue), \qty{10}{\micro\meter} (green), and \qty{15}{\micro\meter} (red) particles in ``reverse wavy'' channels. Flow rates were \qty{197.60}{\micro\litre\per\minute}. a) Experimental results from literature for a reverse wavy channel under similar conditions. Adapted from "A high-throughput micropump with a wide range of flow rates" by \citet{Zhou2018} under CC BY 4.0 (\url{https://creativecommons.org/licenses/by/4.0/}). Image is a cropped version of the original. b) Channel dimensions. c) Simulated results using our geometry-free lift force model.}
    \end{figure}

    \begin{figure}[h]
    \centering
        \includegraphics[width=0.8\textwidth]{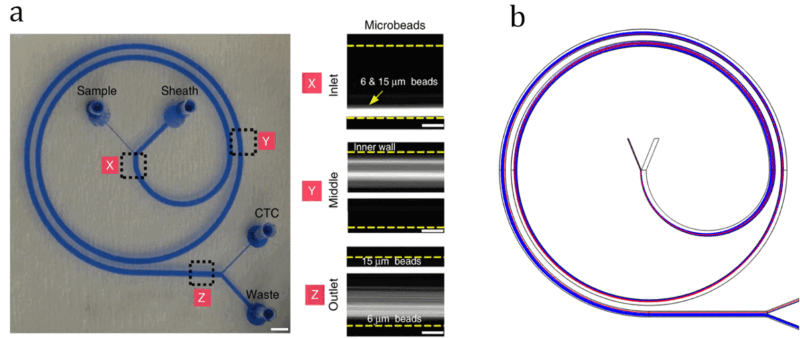}
    \caption[Semicircular channel simulations]{Inertial separation of particles in spiral microchannels. a) Experimental results using \qty{6}{\micro\meter} and \qty{15}{\micro\meter} styrofoam beads. Adapted from \citet{Warkiani2016} with permission from Springer Nature. Scale bars are \qty{2}{\centi\meter}. b) Simulated results using our geometry-free lift force model. Flow rates were \qty{100}{\micro\litre\per\minute}.}
    \end{figure}

\end{appendix}
\end{document}